\documentclass{article}

\PassOptionsToPackage{authoryear, sort&compress}{natbib}

\usepackage[final]{neurips_2024}

\newif\ificml
\icmlfalse

\usepackage{wrapfig}
\usepackage{enumitem}
\usepackage{makecell}
\usepackage[utf8]{inputenc}
\usepackage[T1]{fontenc}
\usepackage{url}
\usepackage{booktabs}
\usepackage{amsfonts}
\usepackage{nicefrac}
\usepackage{xcolor}
\usepackage{colortbl}
\usepackage{microtype}
\usepackage{graphicx}
\usepackage{subcaption}
\usepackage{listings}
\usepackage{fancyvrb}
\usepackage{longtable}
\usepackage{array}
\usepackage[breakable,skins]{tcolorbox}
\usepackage{multirow}
\usepackage{fontawesome5}
\usepackage{caption}

\definecolor{titleblue}{RGB}{20, 60, 120}
\definecolor{linkblue}{RGB}{40, 100, 180}
\definecolor{iconred}{RGB}{190, 55, 55}
\definecolor{iconpurple}{RGB}{110, 70, 150}
\definecolor{iconblue}{RGB}{50, 90, 170}
\definecolor{boxblue}{RGB}{232, 243, 255}  
\definecolor{boxtop}{RGB}{248, 248, 250}
\definecolor{boxbottom}{RGB}{238, 238, 242}
\definecolor{accentblue}{RGB}{60, 120, 200}
\definecolor{abstractbg}{RGB}{245, 248, 255}
\definecolor{abstractbar}{RGB}{60, 120, 200}
\definecolor{sectioncolor}{RGB}{20, 60, 120}
\definecolor{captioncolor}{RGB}{60, 60, 60}

\definecolor{softblue}{RGB}{100, 130, 230}
\definecolor{softcoral}{RGB}{200, 100, 110}
\definecolor{softred}{RGB}{200, 90, 90}
\definecolor{softgreen}{RGB}{90, 160, 110}
\definecolor{impgreen}{RGB}{60, 140, 90}
\definecolor{bestcolor}{RGB}{183, 223, 235}
\definecolor{secondcolor}{RGB}{225, 242, 248}
\definecolor{offlinecolor}{RGB}{232, 245, 233}
\definecolor{onlinecolor}{RGB}{255, 243, 224}

\newtcolorbox{titlebox}{
    enhanced,
    colback=boxtop,
    colframe=boxbottom,
    arc=6pt,
    boxrule=0.4pt,
    left=14pt,
    right=14pt,
    top=14pt,
    bottom=12pt,
    width=\textwidth,
    shadow={0pt}{-1pt}{2pt}{black!8},
    interior style={
        top color=boxtop,
        bottom color=boxbottom,
    },
}

\newcommand{\customtitle}[1]{%
    \begin{center}
    {\fontsize{18}{23}\selectfont\bfseries\color{titleblue}#1}
    \end{center}
    \vspace{0.25cm}
}

\newcommand{\eqcontrib}{$\ast$}

\newcommand{\customauthors}[1]{%
    \begin{center}
    {\large #1}
    \end{center}
    \vspace{0.1cm}
}

\newcommand{\customaffiliations}[1]{%
    \begin{center}
    \footnotesize\itshape
    #1
    \end{center}
}

\newenvironment{customabstract}{%
    \vspace{0.3cm}
    \begin{tcolorbox}[
        enhanced,
        colback=abstractbg,
        colframe=abstractbg,
        boxrule=0pt,
        arc=3pt,
        left=12pt, right=10pt, top=8pt, bottom=8pt,
        borderline west={3pt}{0pt}{abstractbar},
    ]
    \noindent{\bfseries\color{titleblue}Abstract\hspace{0.6em}}%
    \small
}{%
    \end{tcolorbox}
    \vspace{0.1cm}
}

\newcommand{\keywords}[1]{%
    \vspace{0.1cm}
    \noindent{\small\bfseries\color{titleblue}Keywords:}~~{\small\itshape #1}
    \vspace{0.3cm}
}

\newcommand{\paperdate}[1]{%
    \noindent{\small\textcolor{iconred}{\faIcon{calendar-alt}}~\textbf{Date:}~#1}
}

\newcommand{\githubrepo}[1]{%
    \noindent{\small\textcolor{iconpurple}{\faIcon{github}}~\textbf{Github Repo:}~\textcolor{linkblue}{\url{#1}}}
}

\newcommand{\homepage}[1]{%
    \noindent{\small\textcolor{iconblue}{\faIcon{home}}~\textbf{Homepage:}~\textcolor{linkblue}{\url{#1}}}
}

\newcommand{\contactemail}[1]{%
    \noindent{\small\textcolor{iconblue}{\faIcon{envelope}}~\textbf{Contact:}~~#1}
}

\usepackage[
    bookmarks=true,
    colorlinks=true,
    linkcolor=titleblue,
    urlcolor=linkblue,
    citecolor=accentblue,
]{hyperref}

\makeatletter
\let\oldsection\section
\renewcommand{\section}{%
  \@ifstar{\oldsection*}{\@dblarg\colored@section}%
}
\def\colored@section[#1]#2{%
  \oldsection[#1]{\color{sectioncolor}#2}%
}
\let\oldsubsection\subsection
\renewcommand{\subsection}{%
  \@ifstar{\oldsubsection*}{\@dblarg\colored@subsection}%
}
\def\colored@subsection[#1]#2{%
  \oldsubsection[#1]{\color{sectioncolor}#2}%
}
\let\oldsubsubsection\subsubsection
\renewcommand{\subsubsection}{%
  \@ifstar{\oldsubsubsection*}{\@dblarg\colored@subsubsection}%
}
\def\colored@subsubsection[#1]#2{%
  \oldsubsubsection[#1]{\color{sectioncolor}#2}%
}
\makeatother

\usepackage{amsmath}
\usepackage{amssymb}
\usepackage{mathtools}
\usepackage{amsthm}
\theoremstyle{plain}

\theoremstyle{definition}

\theoremstyle{remark}

\usepackage{bm}
\usepackage{algorithm, algorithmic}
\usepackage{tikz}

\usepackage{etoolbox}
\usepackage{natbib}
\newcounter{bibcount}
\makeatletter
\patchcmd{\@lbibitem}{\item[}{\item[\hfil\hspace{2.5em}\stepcounter{bibcount}{[\thebibcount]}\hspace{0.em}}{}{}
\makeatother
\usepackage[nameinlink,capitalise]{cleveref}
\usepackage{titletoc}
\crefname{section}{\S}{\S\S}
\Crefname{section}{\S}{\S\S}

\usepackage{pifont}

\newcommand{\best}[1]{\colorbox{bestcolor}{\textbf{#1}}}

\newcommand{\second}[1]{\colorbox{secondcolor}{\underline{#1}}}

\begin{document}

\thispagestyle{empty}

\begin{titlebox}
\customtitle{Reading $\neq$ Seeing: Diagnosing and Closing the Typography Gap in Vision-Language Models}

\customauthors{%
\textbf{Heng Zhou}\eqcontrib,~~%
\textbf{Ao Yu}\eqcontrib,~~%
\textbf{Li Kang},~~%
\textbf{Yuchen Fan}\\[0.15em]
\textbf{Yutao Fan},~~%
\textbf{Xiufeng Song},~~%
\textbf{Hejia Geng},~~%
\textbf{Yiran Qin}
}

\customaffiliations{%
\eqcontrib Equal contribution
}

\begin{customabstract}
Vision-Language Models achieve near-perfect accuracy at reading text in images, yet prove largely typography-blind: capable of recognizing what text says, but not how it looks. We systematically investigate this gap by evaluating font family, size, style, and color recognition across 26 fonts, four scripts, and three difficulty levels. Our evaluation of 15 state-of-the-art VLMs reveals a striking perception hierarchy: color recognition is near-perfect, yet font style detection remains universally poor. We further find that model scale fails to predict performance and that accuracy is uniform across difficulty levels, together pointing to a training-data omission rather than a capacity ceiling. LoRA fine-tuning on a small set of synthetic samples substantially improves an open-source model, narrowing the gap to the best closed-source system and surpassing it on font size recognition. Font style alone remains resistant to fine-tuning, suggesting that relational visual reasoning may require architectural innovation beyond current patch-based encoders. We release our evaluation framework, data, and fine-tuning recipe to support progress in closing the typographic gap in vision-language understanding.

\end{customabstract}


\noindent
\begin{minipage}[b]{0.75\textwidth}
\paperdate{March 6, 2026}\\[0.1cm]
\homepage{https://henggg.cn/FontBlind/}\\[0.1cm]
\githubrepo{https://github.com/hengzzzhou/FontBlind}\\[0.1cm]
\contactemail{\href{mailto:hengzzzhou@gmail.com}{\textcolor{linkblue}{hengzzzhou@gmail.com}}}
\end{minipage}%
\hfill
\begin{minipage}[b]{0.2\textwidth}
\raggedleft
\end{minipage}
\end{titlebox}

\vspace{0.3cm}

\section{Introduction}\label{sec:introduction}

\begin{figure}[t]
    \centering
    \includegraphics[width=\textwidth]{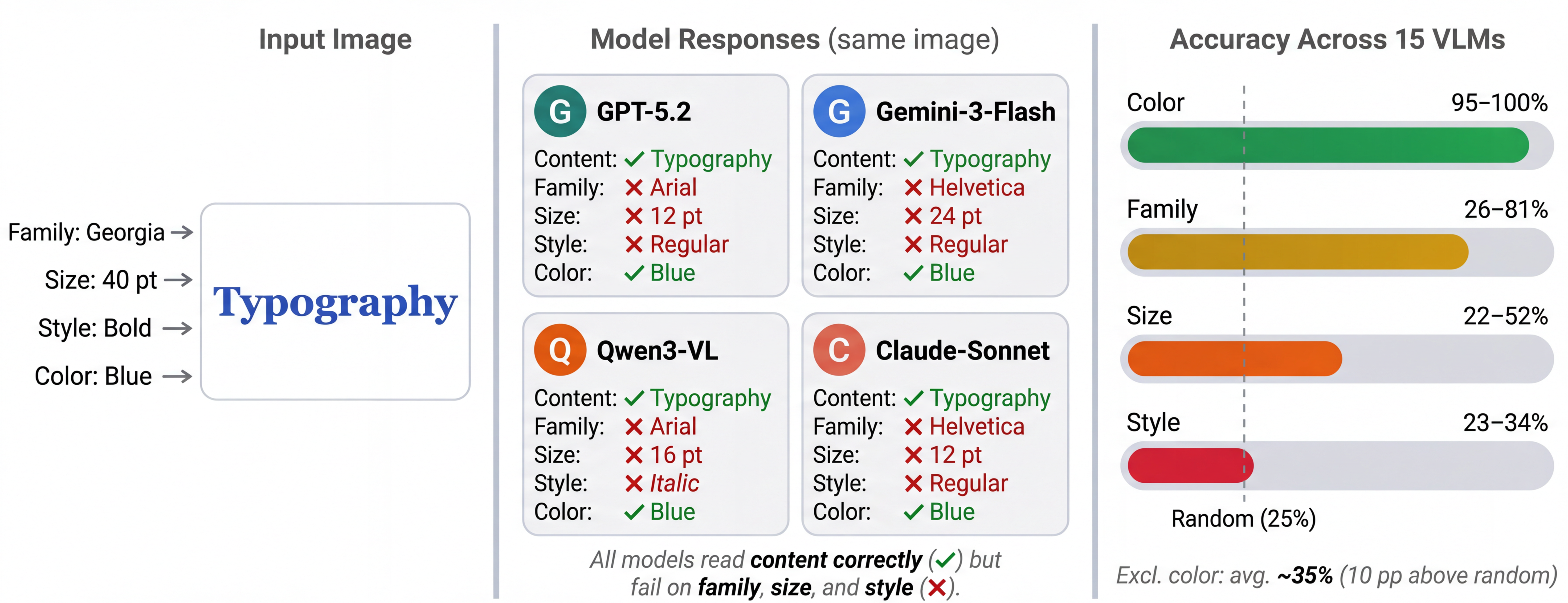}
    \caption{\textbf{Reading $\neq$ Seeing.} Given a single rendered image, four state-of-the-art VLMs unanimously read the text content correctly yet systematically misidentify its font family, size, and style. Only color, a pixel-level cue, is recognized correctly by all models. The right panel quantifies this gap: color accuracy reaches 95--100\%, while font style peaks at only 23--34\%, barely above the 25\% random baseline.}
    \label{fig:teaser}
    \vspace{-1.5em}
\end{figure}

Consider a simple experiment: show a Vision-Language Model an image of the words ``Hello World'' rendered in Times New Roman, 24pt, bold, red, and ask it to describe what it sees. State-of-the-art VLMs will near-perfectly transcribe the text content. But ask ``what font is this?'', ``is it bold?'', or ``how large is the text?'', and the same models stumble dramatically. They can read the text but cannot see how it is rendered, as illustrated in Figure~\ref{fig:teaser}.

This disconnect reveals a fundamental gap in current VLMs. Typography is the visual voice of written language: font choices establish hierarchies, convey tone, and encode meaning beyond the words themselves~\citep{garfield2012fonts, lupton2010thinking, bringhurst2004elements}. Yet existing VLM benchmarks focus on what is in an image, such as objects, scenes, and text content~\citep{liu2023llava, liu2024llava15, bai2023qwenvl, wang2024qwen2vl, chen2024internvl15, liu2024mmbench}, while neglecting how things look. This limitation has concrete consequences: document analysis systems must distinguish headings from body text by font size; accessibility tools must detect emphasis for screen readers; design verification systems must confirm brand-consistent font usage. Each application depends on typographic perception that current VLMs lack.

To systematically diagnose this problem, we construct \textbf{FontBench}, a controlled evaluation framework that isolates typographic perception from content understanding. FontBench evaluates four fundamental typographic properties, namely \textbf{font family}, \textbf{font size}, \textbf{font style}, and \textbf{font color}, across 26 fonts, four scripts, and three difficulty levels. By using synthetically rendered images with known ground truth, FontBench eliminates confounds from image quality, layout complexity, and background clutter, isolating typographic perception as the sole variable under test.

Our evaluation of 15 state-of-the-art VLMs spanning both open-source and closed-source models reveals that typography blindness is not a monolithic failure but follows a structured perception hierarchy: color recognition is near-perfect, font family and size occupy an intermediate zone, and font style remains universally poor, barely above the random baseline. This hierarchy maps directly to visual feature complexity: color reduces to pixel-level statistics, while style demands relational reasoning about stroke weight and slant relative to an internal norm. Excluding the trivially solved color property, average accuracy drops to 35\%, exposing the true depth of the gap.

Two further observations sharpen the diagnosis. The failure is a training gap, not a capacity limitation: model scale fails to predict performance, as larger models do not consistently outperform smaller ones, and accuracy is flat across difficulty levels, exhibiting a binary ``can perceive / cannot perceive'' pattern consistent with typography never being annotated in pretraining data. Confirming this interpretation, LoRA fine-tuning on just 3{,}000 synthetic samples enables a compact open-source model to approach the best closed-source system, with font size recognition surpassing it substantially. Font style, however, remains resistant to fine-tuning, suggesting that relational visual reasoning may require architectural innovation beyond current patch-based encoders. Cross-benchmark evaluation against FRB~\citep{li2025texture} corroborates these findings on an independent test bed. Our contributions are fourfold:

\noindent$\diamond$ \textbf{Problem identification.} We identify and characterize typography blindness as a systematic failure mode in VLMs, where models cannot perceive how text is rendered despite near-perfect content recognition, and provide a controlled evaluation framework covering four properties across four scripts with difficulty stratification.

\noindent$\diamond$ \textbf{Perception hierarchy.} We reveal a consistent hierarchy (color $\gg$ family $>$ size $>$ style) that maps to visual feature complexity, alongside a scaling paradox and difficulty invariance that together identify training data composition as the primary bottleneck.

\noindent$\diamond$ \textbf{Characterization.} We conduct robustness experiments across noise, blur, compression, and rotation, revealing a capability-fragility trade-off, and validate findings on the independent FRB benchmark~\citep{li2025texture}.

\noindent$\diamond$ \textbf{Mitigation via fine-tuning.} We demonstrate that LoRA fine-tuning on 3{,}000 synthetic samples enables a compact model to surpass the best closed-source model on font size and approach it overall, while revealing that font style remains resistant, pointing toward the need for architectural innovation in how encoders handle relational visual comparisons.

\section{Related Work}\label{sec:related_work}

\subsection{Vision-Language Models and Text Understanding}

Recent VLMs have achieved impressive performance on content-level visual tasks, including object recognition, scene understanding, and text-based question answering. Representative systems include LLaVA~\citep{liu2023llava, liu2024llavanext}, Qwen-VL~\citep{bai2023qwenvl, bai2025qwen25vl}, InternVL~\citep{chen2024internvl15, chen2024internvl25}, and proprietary models~\citep{openai2023gpt4, google2023gemini}, typically built on ViT-based vision encoders~\citep{dosovitskiy2021vit, radford2021clip}. These models are evaluated on general multimodal benchmarks~\citep{liu2024mmbench, fu2023mme, yu2024mmvet} and text-centric evaluations~\citep{fanlfqa}, yet fine-grained visual perception remains systematically understudied. Unlike object-level attributes such as shape and texture~\citep{geirhos2019imagenet}, typographic attributes are relational and rendering-specific: they describe not what content says, but how it is visually expressed. Text-centric benchmarks such as TextVQA~\citep{singh2019towards}, DocVQA~\citep{mathew2021docvqa}, and OCRBench~\citep{liu2024ocrbench} evaluate whether models can read text correctly. Similarly, advances in OCR~\citep{li2023trocr, bautista2022parseq} and document-focused VLMs~\citep{ye2023docowl, liu2024textmonkey} further advance text understanding, but none ask whether models perceive the visual presentation of text. This oversight motivates FontBench.

\subsection{Font Recognition}

Font recognition has a long history in document analysis, progressing from hand-crafted feature methods~\citep{zramdini1998optical, jung2001font} to CNN-based approaches such as DeepFont~\citep{wang2015deepfont} and transfer-learning methods~\citep{wang2020cnnfont}. More recently, FontCLIP~\citep{tatsukawa2024fontclip} leverages CLIP embeddings for multilingual font applications, and generative approaches such as DS-Fusion~\citep{tanveer2023dsfusion} explore typography as a creative medium. However, these systems focus almost exclusively on font family identification for Latin script, treating size, style, and color as secondary concerns. Li et al.~\citep{li2025texture} recently introduced FRB (Font Recognition Benchmark), which evaluates 15 common Latin fonts under a multiple-choice format and reveals a stroop effect~\citep{stroop1935}: models tend to report the text content rather than the rendering font. FontBench builds on this foundation while extending it substantially: we evaluate four typographic properties rather than family alone, cover four scripts rather than only Latin, conduct robustness experiments under resolution scaling and image degradation, and investigate fine-tuning as a mitigation strategy. In \cref{subsec:cross_benchmark}, we present a cross-benchmark evaluation on the FRB format.

\subsection{Visual Understanding Benchmarks}

Existing visual understanding benchmarks span a wide range of tasks, from general VQA~\citep{antol2015vqa, goyal2017making, hudson2019gqa} and multimodal reasoning~\citep{yue2023mmmu, lu2023mathvista, lu2022learn} to fine-grained visual recognition of subtle category distinctions in species, vehicles, and products~\citep{wah2011caltech, krause20133d, maji2013fine}. Typographic perception is itself a form of fine-grained visual recognition: distinguishing Arial from Helvetica demands the same within-category discrimination as distinguishing bird subspecies, yet the typographic dimension is entirely absent from existing VLM benchmarks. FontBench is the first benchmark to evaluate all four typographic properties across multiple scripts with a VLM-focused evaluation protocol.

\section{Evaluation Framework}\label{sec:method}

\begin{figure}[t]
    \centering
    \includegraphics[width=\textwidth]{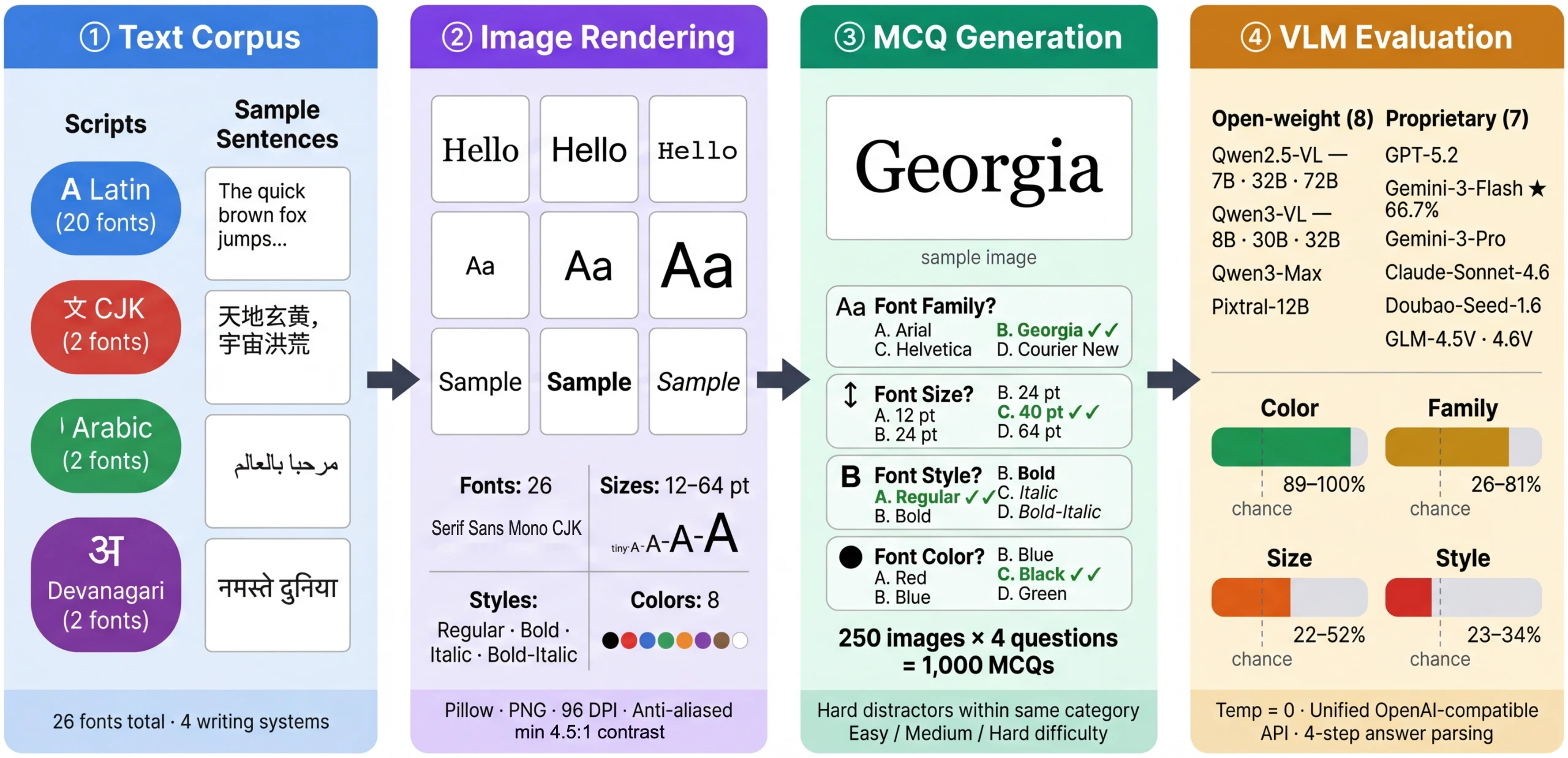}
    \caption{\textbf{Evaluation pipeline.} \textbf{(1) Text Corpus:} script-appropriate sentences spanning Latin, CJK, Arabic, and Devanagari are paired with 26 fonts. \textbf{(2) Image Rendering:} each sentence is rendered at 96\,DPI with anti-aliasing, sampling from 26 fonts $\times$ 8 sizes $\times$ 4 styles $\times$ 8 colors, with background chosen to guarantee a 4.5:1 contrast ratio. \textbf{(3) MCQ Generation:} each rendered image yields four multiple-choice questions, one per property, with hard within-category distractors, producing 1{,}000 questions at three difficulty levels. \textbf{(4) VLM Evaluation:} models are queried at temperature 0; responses are parsed with a 4-step cascade.}
    \label{fig:pipeline}
    \vspace{-1.5em}
\end{figure}

Our central design objective is to isolate typographic perception from content understanding. Each image is synthetically rendered with a single, controlled set of typographic parameters and paired with four multiple-choice questions, one per property, against a 25\% random baseline, as illustrated in Figure~\ref{fig:pipeline}.

\subsection{Font Registry and Data Generation}
\label{subsec:font_registry}

The font registry spans 26 fonts organized by category and script (see \S\ref{app:font_registry}), covering serif, sans-serif, monospace, and display typefaces for Latin, plus dedicated fonts for Chinese, Arabic, and Devanagari. All fonts carry permissive licenses under SIL OFL, Apache, or public domain terms. The registry deliberately includes both highly distinctive typefaces and visually similar ones, so that difficulty levels reflect genuine perceptual discriminability. Representative samples across all four typographic dimensions are shown in Fig.~\ref{fig:gallery}.

Each sample is produced by selecting text from a script-appropriate corpus, sampling typographic parameters according to difficulty stratification rules, and rendering the result using Pillow with anti-aliasing on a background ensuring a minimum contrast ratio of 4.5:1. Difficulty is determined per-property based on visual discriminability. For font family, easy samples pair fonts from distinct categories such as Courier New and Helvetica, while hard samples require distinguishing visually similar typefaces such as Arial and Helvetica Neue. For font size, difficulty scales with the point-size gap between distractors. Style difficulty reflects the perceptual contrast between target and distractors, and color difficulty is governed by chromatic distance. The algorithmic procedure and complete parameter tables are provided in \S\ref{app:generation_details}.

\subsection{Evaluation Protocol}
\label{subsec:questions}

For each sample, we generate four multiple-choice questions, one per property, using varied phrasings to prevent models from exploiting surface-level patterns (see \S\ref{app:questions} for the full template set). Distractor options are drawn from valid values and, for harder questions, are deliberately chosen to be visually similar to the correct answer. Options are randomly ordered to prevent positional bias. All VLMs are queried at temperature 0 for deterministic responses. Answers are parsed with a cascading strategy from exact letter matching to substring search; details are provided in \S\ref{app:implementation}.

\subsection{Dataset Statistics}
\label{subsec:statistics}

\begin{figure}[t]
    \centering
    \includegraphics[width=\textwidth]{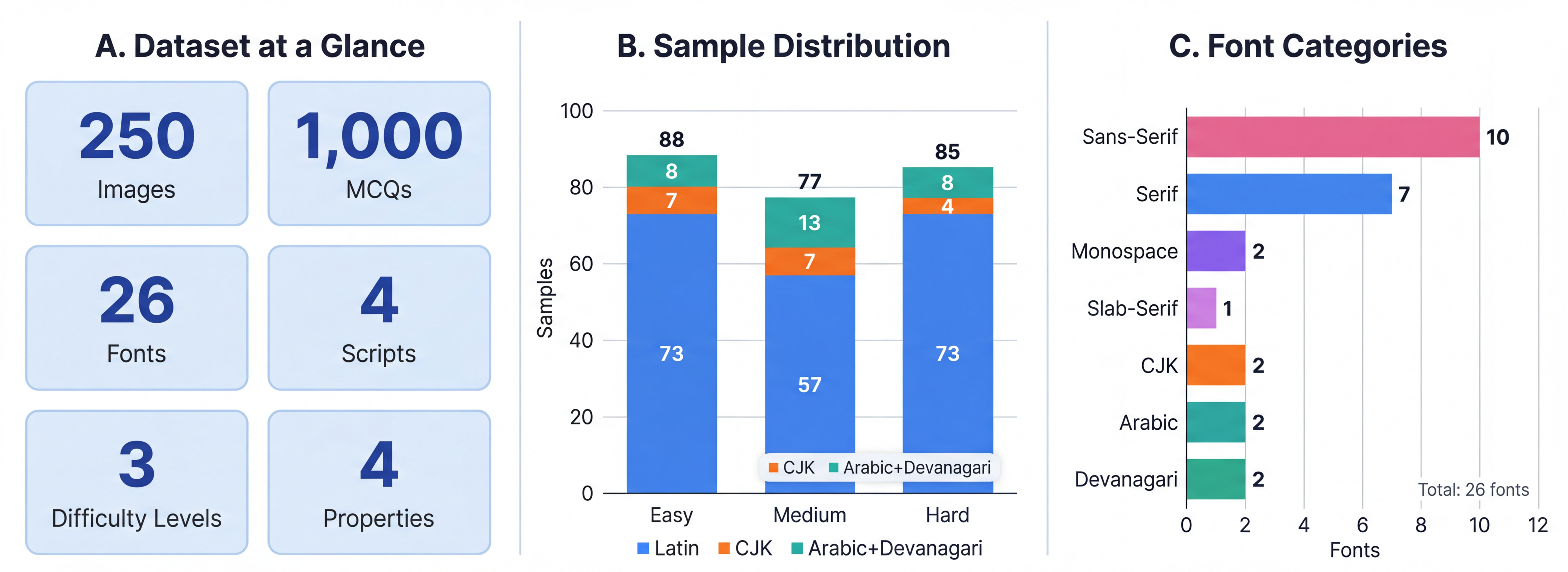}
    \caption{\textbf{Dataset statistics.} \textbf{(A)} Key dimensions. \textbf{(B)} Sample distribution across scripts and difficulty levels. \textbf{(C)} Font category breakdown.}
    \label{fig:dataset_stats}
    \vspace{-1.5em}
\end{figure}

The benchmark comprises 250 samples spanning 26 fonts, 8 sizes, 4 styles, and 8 colors, yielding 1{,}000 multiple-choice questions total. The Latin-heavy distribution at 81.2\% reflects the larger Latin font registry. The three difficulty levels are approximately balanced with Easy at 35.2\%, Medium at 30.8\%, and Hard at 34.0\%. Figure~\ref{fig:dataset_stats} summarizes the composition.

\section{Diagnostic Evaluation}\label{sec:diagnosis}

If VLMs can read text flawlessly, why do they fail to perceive its visual presentation? The answer depends on whether the bottleneck is what models are taught or what architectures can compute. We design our evaluation to distinguish these two hypotheses by probing performance across properties of varying visual complexity, difficulty levels that modulate perceptual challenge, and model scales that test whether capacity alone can compensate. The results point clearly to a training-data omission for most properties, with one notable exception that implicates an architectural limitation.

\subsection{Experimental Setup}
\label{subsec:setup}

We evaluate ten open-source and five closed-source VLMs spanning a wide range of architectures, scales, and training paradigms (see \S\ref{app:model_list} for the full list). Open-source models include the Qwen2.5-VL family~\citep{bai2025qwen25vl} at 7B, 32B, and 72B, Qwen3-VL~\citep{bai2025qwen3vl} at 8B, 30B-A3B (MoE), and 32B, along with Pixtral-12B~\citep{agrawal2024pixtral}, GLM-4.5V and GLM-4.6V~\citep{vteam2026glm45v}. Closed-source models comprise Qwen3-Max, GPT-5.2~\citep{singh2025gpt5}, Gemini-3-Flash and Gemini-3-Pro~\citep{google2023gemini}, Claude-Sonnet-4.6, and Doubao-Seed-1.6. All are queried at temperature 0 with PNG images at native resolution. Random baseline accuracy is 25\%.

\subsection{The Perception Hierarchy}
\label{subsec:hierarchy}

\begin{table*}[t]
    \centering
    \caption{Main results on FontBench. We report accuracy (\%) for each property, overall, by difficulty level, and by writing system. \best{Best} and \second{second best} results are highlighted. Random baseline is 25\%.}
    \label{tab:main_results}
    \resizebox{\textwidth}{!}{%
    \begin{tabular}{@{}l cccc c ccc ccc @{}}
        \toprule
        & \multicolumn{4}{c}{\textbf{Per Property}} & & \multicolumn{3}{c}{\textbf{By Difficulty}} & \multicolumn{3}{c}{\textbf{By Script}} \\
        \cmidrule(lr){2-5} \cmidrule(lr){7-9} \cmidrule(lr){10-12}
        \textbf{Model} & \textbf{Family} & \textbf{Size} & \textbf{Style} & \textbf{Color} & \textbf{Overall} & \textbf{Easy} & \textbf{Med} & \textbf{Hard} & \textbf{Latin} & \textbf{CJK} & \textbf{Other} \\
        \midrule
        \multicolumn{12}{l}{\textit{Open-Source Models}} \\
        Qwen2.5-VL-7B & 35.2 & 44.4 & 27.6 & 97.6 & 51.2 & 51.4 & 53.2 & 49.1 & 49.8 & 65.3 & 52.6 \\
        Qwen2.5-VL-32B & 36.0 & 34.4 & 30.4 & 89.2 & 47.5 & 49.4 & 45.1 & 47.6 & 47.3 & 59.7 & 41.4 \\
        Qwen2.5-VL-72B & 38.8 & 36.0 & \second{33.2} & 96.4 & 51.1 & 49.7 & 52.6 & 51.2 & 50.7 & 65.3 & 44.8 \\
        Qwen3-VL-8B & 36.8 & 39.2 & 28.8 & 99.6 & 51.1 & 50.3 & 53.2 & 50.0 & 49.5 & 66.7 & 52.6 \\
        Qwen3-VL-30B-A3B & 49.6 & 40.4 & 28.0 & 99.6 & 54.4 & 54.3 & 54.9 & 54.1 & 54.8 & 58.3 & 49.1 \\
        Qwen3-VL-32B & 42.4 & 37.2 & 26.0 & \best{100.0} & 51.4 & 51.4 & 52.9 & 50.0 & 50.7 & 66.7 & 46.6 \\
        Pixtral-12B & 26.0 & 27.2 & 28.4 & 24.8 & 26.6 & 25.9 & 26.6 & 27.4 & 27.1 & 26.4 & 23.3 \\
        GLM-4.5V & 25.2 & 22.0 & 22.8 & 26.8 & 24.2 & 22.7 & 24.0 & 25.9 & 23.3 & 29.2 & 27.6 \\
        GLM-4.6V & 39.2 & 35.6 & 25.6 & \best{100.0} & 50.1 & 48.3 & 50.3 & 51.8 & 49.5 & 61.1 & 47.4 \\
        \midrule
        \multicolumn{12}{l}{\textit{Closed-Source Models}} \\
        Qwen3-Max & 46.0 & 32.4 & 31.2 & 99.2 & 52.2 & 51.4 & 52.6 & 52.6 & 51.4 & 65.3 & 50.0 \\
        GPT-5.2 & 58.8 & \second{50.0} & 31.2 & 99.6 & \second{59.9} & \second{61.6} & \second{59.7} & \second{58.2} & 59.5 & 69.4 & \second{56.9} \\
        Gemini-3-Flash & \best{80.8} & \best{52.4} & \best{33.6} & \best{100.0} & \best{66.7} & \best{65.9} & \best{67.2} & \best{67.1} & \best{67.7} & \second{72.2} & 56.0 \\
        Gemini-3-Pro & 40.8 & 41.2 & 32.0 & 94.0 & 52.0 & 50.9 & 53.2 & 52.1 & 51.2 & \second{72.2} & 44.8 \\
        Claude-Sonnet-4.6 & \second{64.0} & 44.8 & 28.0 & 97.6 & 58.6 & 60.2 & 57.8 & 57.6 & \second{60.3} & 62.5 & 44.0 \\
        Doubao-Seed-1.6 & 44.8 & 44.4 & 30.4 & 98.8 & 54.6 & 53.1 & 57.8 & 53.2 & 52.0 & \best{73.6} & \best{61.2} \\
        \midrule
        Random & 25.0 & 25.0 & 25.0 & 25.0 & 25.0 & 25.0 & 25.0 & 25.0 & 25.0 & 25.0 & 25.0 \\
        \bottomrule
    \end{tabular}%
    }
\end{table*}

The most striking pattern in Table~\ref{tab:main_results} is not which model wins, but that all models fail in the same structured way. A universal perception hierarchy emerges: color recognition is near-perfect at 89--100\%, font family and size occupy an intermediate zone of 22--81\%, and font style peaks at only 33.6\%, barely above the 25\% random baseline. This hierarchy holds across architectures, scales, and training paradigms. Even the best model, Gemini-3-Flash at 66.7\%, drops to 55.6\% once color is excluded. For most open-source models, the color-excluded accuracy is approximately 35\%, revealing that the headline numbers substantially mask the true depth of typography blindness.

\textbf{Why this hierarchy?} The ordering is not arbitrary; it maps onto a hierarchy of computational complexity within ViT-based encoders~\citep{dosovitskiy2021vit} (see Fig.~\ref{fig:hierarchy} for an illustration). Color is a zeroth-order feature: the average RGB statistics within a single patch suffice, requiring no spatial reasoning. Font size is first-order: it depends on how many patches the rendered glyphs span, requiring spatial extent estimation. Font family is categorical: it demands matching high-frequency glyph contours against learned prototypes, a task for which ViT patch embeddings provide at least partial signal. Font style, however, is fundamentally relational: bold is defined as thicker strokes relative to regular, italic as a slant relative to upright. The model must implicitly compare the observed stroke weight against an internal norm for that typeface, a second-order comparison that current encoders and training objectives do not explicitly support. This computational analysis predicts that all properties except style should respond to fine-tuning, while style may require architectural changes, a prediction we verify in \cref{sec:remedy}.

\paragraph{Difficulty Invariance: A Diagnostic Signature.}
\label{subsec:difficulty_analysis}

If a model had partially learned typographic features, we would expect graded performance: high accuracy on easy samples such as distinguishing Courier New from Helvetica, degrading on hard ones like Arial vs.\ Helvetica Neue. Instead, most models show remarkably flat accuracy across difficulty levels. Qwen3-VL-8B scores 50.3\% on easy, 53.2\% on medium, and 50.0\% on hard, with medium actually performing best, as shown in Table~\ref{tab:main_results}. This difficulty invariance is a diagnostic signature of missing perceptual features: models that cannot perceive a property at all will guess at random regardless of how visually distinct the options are. The flat profile indicates a binary ``can perceive / cannot perceive'' boundary, precisely the pattern expected from a training data omission rather than weak perceptual capacity. Only Gemini-3-Flash shows the expected difficulty sensitivity, with 65.9\% on easy, 67.2\% on medium, and 67.1\% on hard, confirming that its superior performance stems from genuinely more capable visual feature extraction.

\paragraph{Script-Specific Perception Reflects Training Bias.}
\label{subsec:script_analysis}

Typographic perception is not script-agnostic. Qwen3-VL-8B performs substantially better on CJK at 66.7\% than Latin at 49.5\%, likely reflecting Alibaba's CJK-heavy training corpora, while Doubao-Seed-1.6 leads on Arabic and Devanagari at 61.2\%, suggesting ByteDance's multilingual emphasis. These asymmetries support the training-data hypothesis: models excel on scripts likely overrepresented in their pretraining data. A script-agnostic perceptual capability would show uniform performance across writing systems; the observed biases indicate that typographic signal is acquired incidentally through script-specific data composition rather than deliberate typographic training.

\paragraph{How Models Fail Reveals Why They Fail.}
\label{subsec:property_analysis} Per-property error patterns expose the mechanisms behind typography blindness; detailed breakdowns are in \S\ref{app:property_analysis}. For font family, within-category confusion dominates: models readily distinguish serif from monospace but cannot separate Arial from Helvetica, indicating that they capture coarse category-level features but lack fine-grained glyph discrimination. For font size, models employ fundamentally different estimation strategies: GPT-5.2 succeeds at extremes with 76.3\% on Small and 90.3\% on XLarge but fails on intermediate sizes at 35.5\% on Medium, while Qwen3-VL-8B and Gemini-3-Flash achieve 0\% on XLarge, suggesting they lack the concept of very large text altogether. Most revealing is font style: confusion matrices show a dominant ``regular'' bias where models predict regular for 67--80\% of all samples regardless of true style, with bold-italic essentially never predicted. This is not a classification error but a form of perceptual blindness, confirming that current VLMs are effectively blind to font style variations.

\subsection{The Scaling Paradox: Why More Parameters Do Not Help}
\label{subsec:scaling}

A counterintuitive finding emerges from the scaling analysis, detailed in \S\ref{app:scaling}: bigger models are not better at typography. In the Qwen2.5 family, the 7B model at 51.2\% outperforms the 32B at 47.5\%, while the 72B recovers only to 51.1\%. In the Qwen3 family, the 30B-A3B MoE variant at 54.4\% outperforms the 32B dense model at 51.4\%. This paradox has a clear implication: typography blindness is not a capacity limitation that more parameters can overcome. The relevant visual features are either present or absent in the training data; adding parameters to a model that has never seen typographic annotations does not create typographic perception. This motivates a targeted intervention through fine-tuning, which we investigate in \cref{sec:remedy}.

\section{Robustness and External Validation}\label{sec:analysis}

The perception hierarchy and scaling paradox were established under controlled conditions. Two questions remain: (1) do these findings survive real-world image degradation, and (2) do they generalize to independently designed benchmarks? The answers reveal a new insight, a capability-fragility trade-off, and confirm our findings on an external test bed. We further probe the mechanistic source of failures through attention analysis.

\subsection{Resolution Sensitivity: Models Are Calibrated, Not Generalizable}
\label{subsec:resolution}

In deployment, document images arrive at arbitrary scales. We evaluate three representative models at four resolutions (0.25$\times$, 0.5$\times$, 1$\times$, 2$\times$).\footnote{Baseline accuracies reported in this section correspond to robustness-set evaluations and may differ slightly from Table~\ref{tab:main_results} owing to API response variability across evaluation runs.} The key finding is that the resolution--accuracy relationship is non-monotonic (Figure~\ref{fig:res_rob}a): GPT-5.2 and Gemini-3-Flash both peak at the original 1$\times$ resolution and degrade in both directions, upscaling hurts GPT-5.2 by 4.4pp, while downscaling to 0.25$\times$ costs both models over 12pp. This non-monotonicity reveals that these models are implicitly calibrated to the resolution range of their training data rather than possessing resolution-invariant typographic perception. In contrast, Qwen3-VL-8B remains stable across the entire 4$\times$ range (49.7--51.9\%), suggesting that its weaker but more robust perception relies on structural features rather than resolution-sensitive pixel patterns.

The property-level effects are also revealing. Font size recognition collapses to the 25\% random baseline at 0.25$\times$ for both GPT-5.2 and Gemini-3-Flash, and GPT-5.2 also collapses at 2$\times$, indicating it is narrowly tuned to a specific resolution window for size estimation. Font family improves at 2$\times$ as finer glyph details become visible, while color remains robust above 97\% at all resolutions, consistent with color being a low-frequency, resolution-invariant signal.

\subsection{The Capability-Fragility Trade-off}
\label{subsec:robustness}

\begin{figure}[t]
    \centering
    \includegraphics[width=\textwidth]{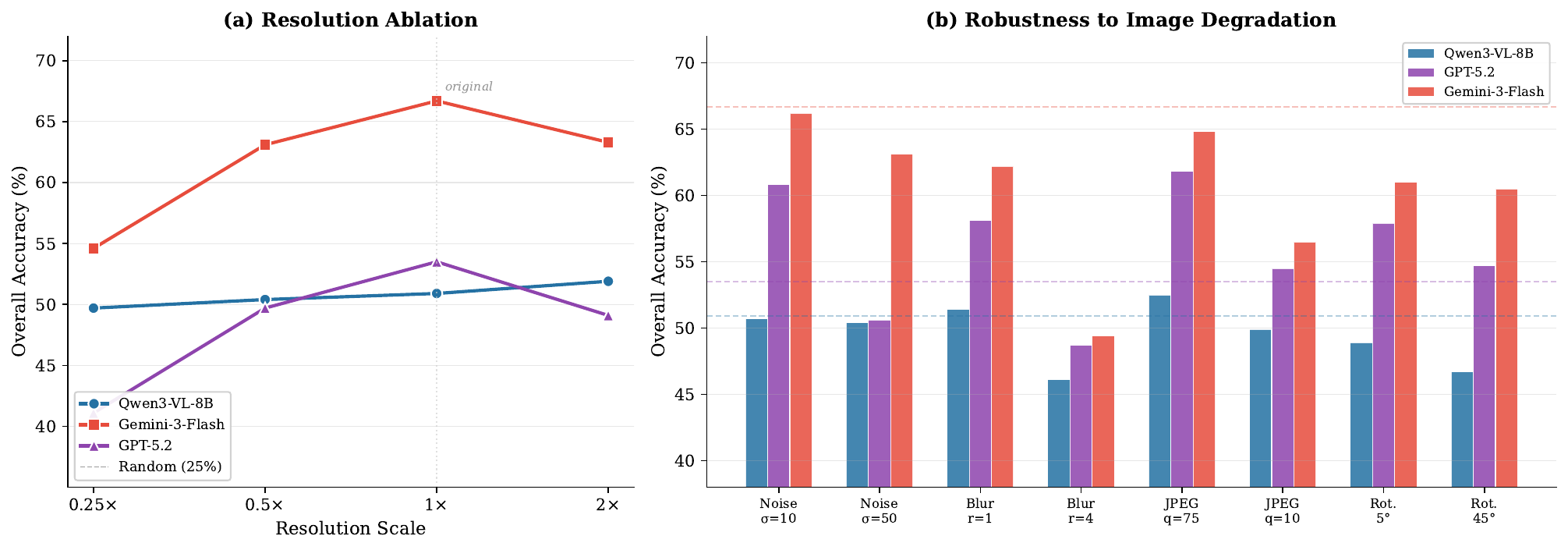}
    \caption{(a)~Resolution ablation: GPT-5.2 and Gemini-3-Flash peak at 1$\times$ with non-monotonic degradation in both directions; Qwen3-VL-8B remains stable. (b)~Robustness to image degradation: Gemini-3-Flash suffers the largest absolute drops under blur ($-$17.3pp) and JPEG compression ($-$10.2pp), while Qwen3-VL-8B is the most stable. Dashed lines indicate 1$\times$ baselines.}
    \label{fig:res_rob}
    \vspace{-1.5em}
\end{figure}

We stress-test three models under Gaussian noise ($\sigma \in \{10, 50\}$), blur ($r \in \{1, 4\}$), JPEG compression ($q \in \{75, 10\}$), and rotation ($\theta \in \{5^\circ, 45^\circ\}$). The central finding, illustrated in Figure~\ref{fig:res_rob}b, is a capability-fragility trade-off: the best clean-image model is the most vulnerable to degradation.

Gemini-3-Flash, despite leading on clean images at 66.7\%, drops by 17.3pp under blur and 10.2pp under JPEG compression. Qwen3-VL-8B, by contrast, maintains 46--53\% across all conditions (6.4pp total range). This divergence reveals that high-performing models rely on brittle high-frequency glyph details that blur and compression destroy, while weaker models depend on coarser structural features that survive degradation. The practical implication is significant: for real-world document analysis where images may be scanned, compressed, or photographed at angles, the best benchmark model may not be the most reliable. Interestingly, mild JPEG compression improves GPT-5.2 (53.5\% $\to$ 61.8\%), suggesting that light compression smooths fine-grained noise that otherwise disrupts font perception. Color recognition remains robust above 90\% even under severe degradations, further confirming that color perception relies on low-frequency signals immune to high-frequency corruption.

\subsection{Cross-Benchmark Validation: Is This Real or Benchmark-Specific?}
\label{subsec:cross_benchmark}

A critical validity question is whether FontBench's findings reflect genuine perceptual gaps or benchmark-specific artifacts. We evaluate all models on FRB~\citep{li2025texture}, an independently designed benchmark that tests 15 Latin fonts under two conditions: normal sentences (easy) and text that spells a conflicting font name (hard, creating a ``stroop effect''~\citep{stroop1935}). Full per-model FRB results are provided in \S\ref{app:frb_full}; fine-tuning transfer results are included in Table~\ref{tab:finetuning_results}.

Despite FRB's much harder 15-way classification with a 6.7\% random baseline compared to FontBench's 25\%, model rankings are strikingly consistent: Gemini-3-Flash dominates at 40.5\%, followed by Claude-Sonnet-4.6 at 22.9\% and GPT-5.2 at 20.8\%. This rank consistency across two independently designed benchmarks confirms that the measured differences reflect genuine perceptual capability rather than format-specific shortcuts.

The stroop effect reveals an even deeper insight about how VLMs process font-related queries. When text spells a conflicting font name, all Qwen models collapse to the 6.7\% random baseline, as shown in Fig.~\ref{fig:stroop}, indicating that text content fully overrides visual font perception. This suggests that font queries are routed through the language pathway rather than the visual encoder: the model reads ``Times New Roman'' and outputs that as its answer, regardless of the actual rendered font. This OCR-dominance mechanism explains why typography blindness persists despite strong visual encoders. The scaling paradox also replicates on FRB, where the Qwen3 family scores 9.6\% at 8B, 9.9\% at 30B-A3B, and 10.1\% at Max, confirming its robustness as a general phenomenon beyond any single benchmark.

\subsection{Attention Analysis: Where Models Look When They Fail}
\label{subsec:attention}

\begin{figure}[t]
    \centering
    \includegraphics[width=\textwidth]{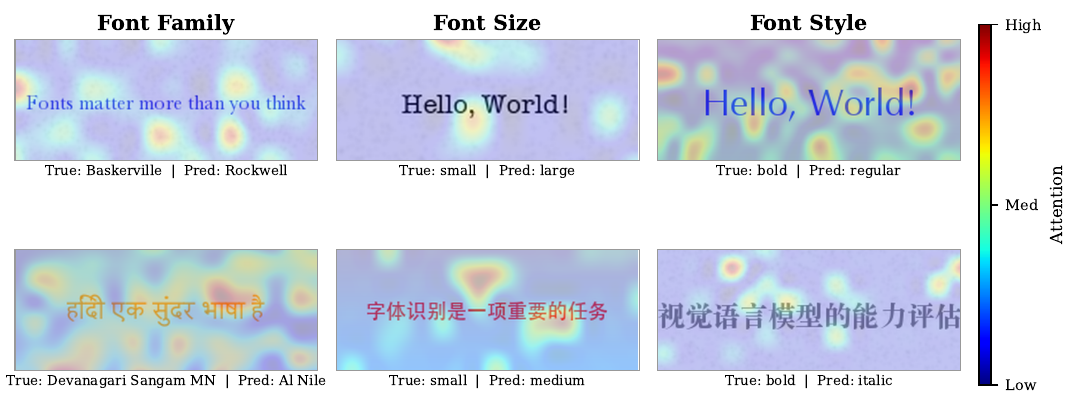}
    \caption{\textbf{Attention heatmaps for hard-split failures.} Self-attention weights from the final decoder layer of Qwen2.5-VL-7B, averaged over all heads. \textbf{Font Family:} attention spreads uniformly; discriminative glyph features are not attended to. \textbf{Font Size:} attention covers only part of the text, failing to integrate spatial extent. \textbf{Font Style:} attention is character-level but assigns equal weight to all strokes, missing the relative thickness that distinguishes bold from regular.}
    \label{fig:attention}
    \vspace{-1.5em}
\end{figure}

To understand why the perception hierarchy emerges, we visualize self-attention weights from the final decoder layer of Qwen2.5-VL-7B, averaged across all heads, for six hard-split samples that the model answers incorrectly. As shown in Figure~\ref{fig:attention}, three qualitatively distinct failure modes emerge.

For \textbf{font family}, attention is diffuse and nearly uniform across all character positions rather than focusing on discriminative glyph-level features such as serif terminals and stroke modulation. For \textbf{font size}, attention is spatially truncated, covering only part of the text while ignoring the rest, preventing the spatial extent integration that size estimation requires. For \textbf{font style}, attention is character-level but assigns nearly equal weight to all strokes. Distinguishing bold from regular requires a relative comparison of stroke thickness to a norm, and the attention maps show no evidence of such differential weighting. This explains why style is the most resistant property even after fine-tuning: the failure is not about where the model looks but about what it computes from what it sees.

\section{Closing the Gap: Fine-Tuning for Typographic Perception}\label{sec:remedy}

The preceding analysis identifies training data composition as the primary bottleneck and predicts that font family, size, and color should respond to targeted training, while font style may resist due to its relational nature. We now test this prediction through targeted LoRA fine-tuning experiments.

\subsection{Setup and Training Protocol}
\label{subsec:ft_setup}

We generate a dedicated fine-tuning dataset of 3{,}000 image-question-answer triplets using the same synthetic pipeline described in \cref{sec:method}, covering all four font properties with balanced distributions. The training images use different random seeds and text samples than the evaluation benchmark to ensure no data leakage. We fine-tune three model scales: Qwen2.5-VL-7B with 7.6B parameters, Qwen3-VL-8B with 8.3B, and Qwen2.5-VL-32B with 32.8B. The two mid-scale models provide a clean test of fine-tuning efficacy at full precision; the 32B model explores larger-scale fine-tuning under the constraint that it must use 4-bit NF4 quantization to fit on a single GPU, which introduces a confound noted below.

We apply LoRA adapters~\citep{hu2021lora, dettmers2023qlora, zhang2023llamaadapter} with rank $r{=}16$, $\alpha{=}32$, and dropout 0.05 to the \texttt{q/k/v/o\_proj} attention matrices. Training uses AdamW with learning rate $2{\times}10^{-4}$, warmup ratio 0.05, batch size 1 with 16 gradient accumulation steps, BF16 precision for the 7B and 8B models, and runs for 3 epochs. Each adapter trains only 0.2--0.3\% of total parameters. Full hyperparameter details are provided in \S\ref{app:finetuning}.

\subsection{Results}
\label{subsec:ft_results}

\begin{table}[t]
    \centering
    \caption{Fine-tuning results on FontBench (4-way MCQ, 25\% baseline) and cross-benchmark transfer on FRB~\citep{li2025texture} (15-way MCQ, 6.7\% baseline). \best{Best} results highlighted. Full FRB results for all models are in \S\ref{app:frb_full}.}
    \label{tab:finetuning_results}
    \begin{tabular}{@{}lccccc|c@{}}
        \toprule
        & \multicolumn{5}{c}{\textbf{FontBench}} & \textbf{FRB} \\
        \cmidrule(lr){2-6} \cmidrule(lr){7-7}
        \textbf{Model} & \textbf{Family} & \textbf{Size} & \textbf{Style} & \textbf{Color} & \textbf{Overall} & \textbf{Overall} \\
        \midrule
        \multicolumn{7}{l}{\textit{Baseline Models, Zero-Shot}} \\
        Qwen2.5-VL-7B & 35.2 & 44.4 & 27.6 & 97.6 & 51.2 & 6.9 \\
        Qwen3-VL-8B & 36.8 & 39.2 & 28.8 & 99.6 & 51.1 & 9.6 \\
        \midrule
        \multicolumn{7}{l}{\textit{Fine-Tuned Models, LoRA}} \\
        Qwen2.5-VL-7B + LoRA & 46.0 & \best{66.4} & 27.6 & 98.0 & 59.5 & 9.6 \\
        Qwen3-VL-8B + LoRA & \best{52.8} & 60.0 & \best{30.4} & \best{100.0} & \best{60.8} & 13.1 \\
        Qwen2.5-VL-32B + LoRA & 43.6 & 34.4 & 31.2 & 94.8 & 51.0 & \best{13.3} \\
        \bottomrule
    \end{tabular}
\end{table}

The results in Table~\ref{tab:finetuning_results} confirm the prediction with striking precision. LoRA fine-tuning on just 3{,}000 synthetic samples lifts Qwen3-VL-8B from 51.1\% to 60.8\% and Qwen2.5-VL-7B from 51.2\% to 59.5\%, narrowing the gap to Gemini-3-Flash from 15.6pp to 5.9pp. The 32B model improves only modestly by 3.5pp, though this is confounded by 4-bit NF4 quantization required to fit on a single GPU.

Crucially, the gains follow a learnability hierarchy that mirrors the perception hierarchy. Font size benefits most dramatically: Qwen2.5-VL-7B + LoRA achieves 66.4\%, a 22.0pp gain that surpasses Gemini-3-Flash by 14pp. Font family improves substantially by 16.0pp, and color reaches a perfect 100\%. Font style, as predicted by our architectural analysis, is conspicuously resistant: gains are at most 1.6pp across all model sizes. This selective response to fine-tuning is the strongest evidence for our two-factor explanation: family, size, and color fail because the relevant visual features are already latent in the ViT encoder but never activated by training signal; style fails because the encoder lacks the computational primitives for relational comparison.

\begin{figure}[t]
    \centering
    \includegraphics[width=\textwidth]{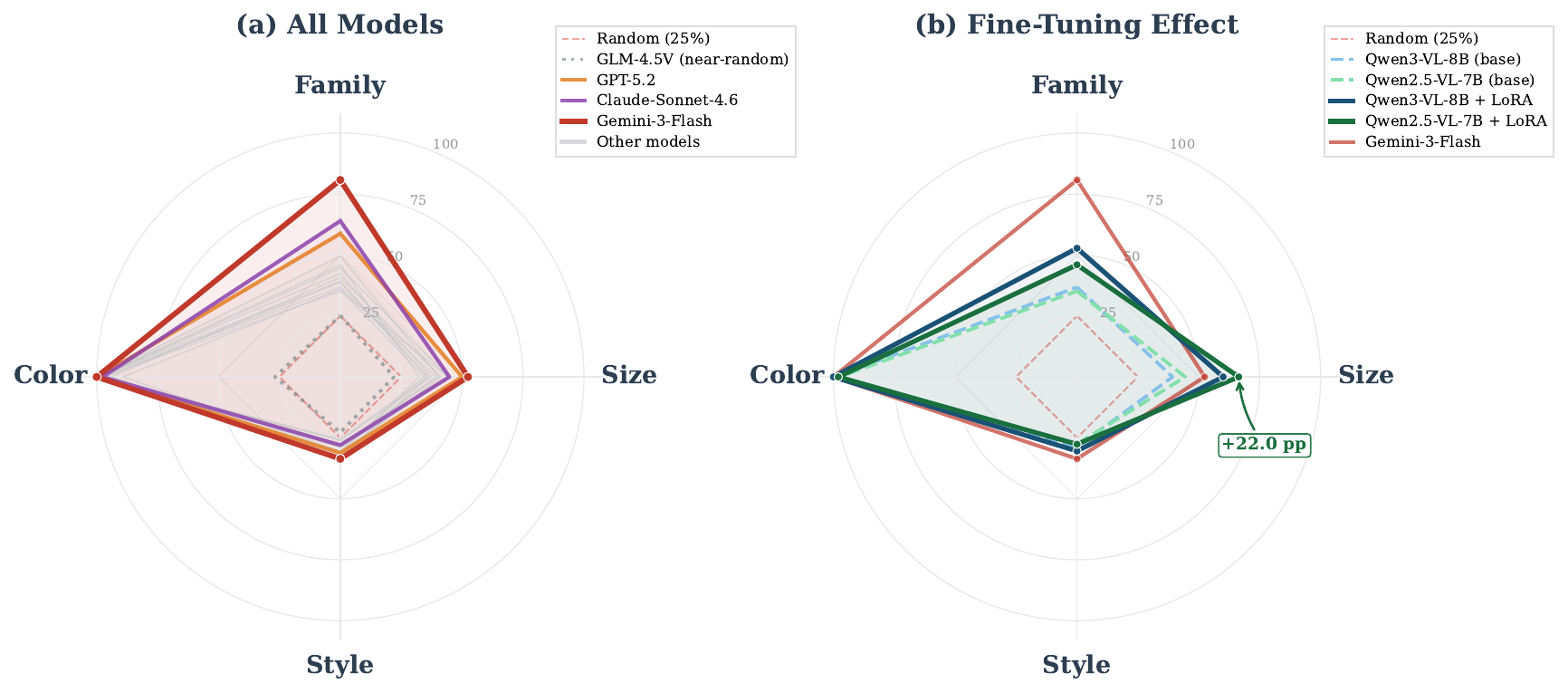}
    \caption{Radar charts of per-property accuracy. (a)~All models share a ``spiky'' pattern with color $\gg$ others; Gemini-3-Flash dominates while GLM-4.5V collapses. (b)~LoRA fine-tuning reshapes the capability profile: font size surpasses even Gemini-3-Flash after a 22.0pp gain for the 7B model, while the overall gap narrows to 5.9pp.}
    \label{fig:radar}
    \vspace{-1.5em}
\end{figure}

\subsection{What Fine-Tuning Reveals About the Failure}
\label{subsec:ft_implications}

Fine-tuning serves not only as a remedy but as a diagnostic instrument. The selective response across properties answers the training gap vs.\ architectural limitation question with precision:

\textbf{Latent features, missing supervision.} For family, size, and color, the fact that 3{,}000 samples suffice for large gains indicates that the ViT encoder already computes the relevant visual features. These features are latent in the representation but never connected to typographic labels during pretraining. Fine-tuning does not teach the model to see new things; it teaches the model to report what it already sees.

\textbf{Missing computational primitives.} Font style's resistance across all model sizes, with gains of only 0.0--1.6pp, points to a qualitatively different bottleneck. Style perception requires comparing the observed stroke weight against an internal norm for each typeface, a relational computation that current patch-based encoders do not natively support. Architectural innovations such as contrastive modules or explicit style-comparison heads may be necessary.

\textbf{OCR dominance survives fine-tuning.} Despite improved FRB-easy performance where Qwen3-VL-8B + LoRA improves from 14.0\% to 22.7\%, all fine-tuned models remain at the 6.7\% random baseline on FRB's stroop set, as discussed in \cref{subsec:cross_benchmark}. Fine-tuning teaches the model to recognize fonts when text is neutral but cannot override the deeper tendency to route font queries through the language pathway when text content conflicts with visual evidence. Mitigating this OCR-dominance mechanism likely requires training on conflicting text-font pairs or modifications to the visual encoding pathway.

In summary, typography blindness has two distinct causes that demand different solutions. For three of four properties, the fix is straightforward: modest synthetic data activates latent visual features, enabling a compact open-source model to approach or surpass the best closed-source system. For font style, closing the gap requires fundamentally rethinking how vision encoders represent and compare relational visual properties.

\section{Conclusion}\label{sec:conclusion}

Our work reveals that VLMs suffer from a systematic disconnect between reading and seeing: they can transcribe text near-perfectly but remain largely blind to how it is visually rendered. The failure is not monolithic but structured, following a perception hierarchy that maps onto computational complexity in ViT encoders. Three converging lines of evidence, difficulty invariance, the scaling paradox, and selective fine-tuning response, establish that the primary cause is a training data omission: typographic properties are never annotated in pretraining data, leaving latent visual features unactivated. For font family, size, and color, some samples suffice to activate these features and close much of the gap. Font style is qualitatively different: its relational nature resists fine-tuning across all model sizes, pointing to a need for architectural innovation in how vision encoders represent relative visual attributes. Attention analysis corroborates this mechanistically, and cross-benchmark validation on FRB confirms these findings generalize beyond FontBench. Extending the evaluation to real-world document images with mixed fonts and complex layouts remains an important direction for future work. We release our evaluation framework, data, and fine-tuning recipe to support progress toward closing the typographic gap in vision-language understanding.

\clearpage

\bibliography{main}

\clearpage

\appendix

\vbox{%
  \hsize\textwidth
  \linewidth\hsize
  \vskip 0.1in
  \hrule height 4pt
  \vskip 0.25in
  \vskip -\parskip
  \centering
  {\LARGE\bf\color{titleblue} Reading $\neq$ Seeing: Diagnosing and Closing the Typography Gap in Vision-Language Models\\[0.15cm](Appendix)}
  \vskip 0.29in
  \vskip -\parskip
  \hrule height 1pt
  \vskip 0.09in
}

\startcontents[appendix]
\vspace{0.5cm}
\printcontents[appendix]{}{1}{\setcounter{tocdepth}{2}}
\vspace{1cm}

\section{Benchmark Construction}
\label{app:benchmark}

\subsection{Font Registry and Sample Gallery}
\label{app:font_registry}

\begin{table}[h]
    \centering
    \caption{Font categories in FontBench. The registry spans multiple typeface classifications and four writing systems.}
    \label{tab:font_categories}
    \begin{tabular}{@{}llp{8cm}@{}}
        \toprule
        \textbf{Category} & \textbf{Script} & \textbf{Fonts: 20 Latin, 2 CJK, 4 Other} \\
        \midrule
        Serif & Latin & Times New Roman, Georgia, Palatino, Baskerville, Didot, Cochin, Rockwell \\
        Sans-Serif & Latin & Arial, Arial Bold, Helvetica, Helvetica Neue, Verdana, Futura, Avenir, Gill Sans, Optima, Trebuchet MS \\
        Monospace & Latin & Courier New, Menlo \\
        Display & Latin & American Typewriter \\
        \midrule
        CJK & Chinese & STHeiti, Songti SC \\
        \midrule
        Arabic & Arabic & Al Nile, Baghdad \\
        Devanagari & Devanagari & Devanagari MT, Devanagari Sangam MN \\
        \bottomrule
    \end{tabular}
\end{table}

\begin{figure}[h]
    \centering
    \includegraphics[width=\textwidth]{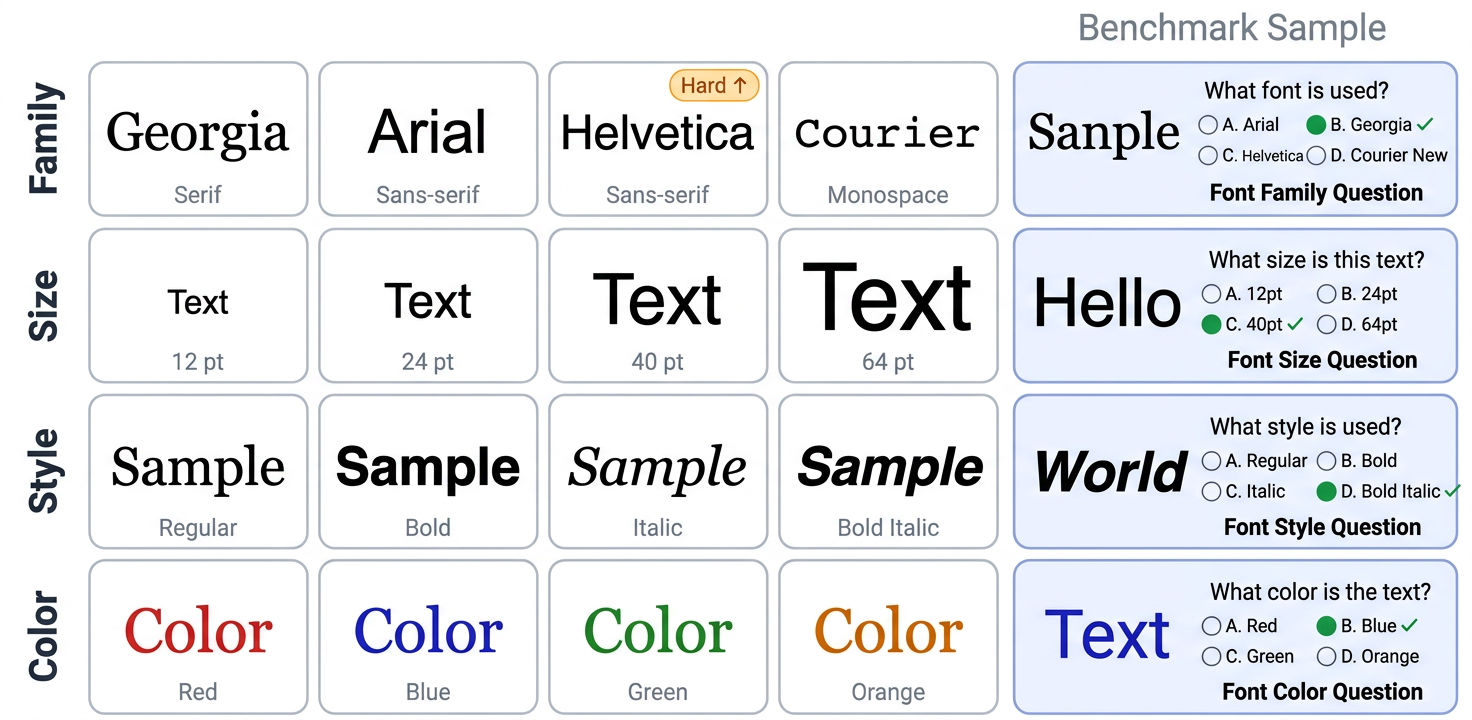}
    \caption{\textbf{FontBench sample gallery.} The four typographic dimensions evaluated: font family (note Arial vs.\ Helvetica are visually near-identical), font size (12--64pt), font style (regular/bold/italic/bold-italic), and font color. The rightmost column shows a representative question for each dimension.}
    \label{fig:gallery}
\end{figure}

\subsection{Data Generation Details}
\label{app:generation_details}

\subsubsection{Generation Algorithm}

Algorithm~\ref{alg:generation} formalizes the synthetic sample generation procedure.

\begin{algorithm}[h]
    \caption{Synthetic Sample Generation}
    \label{alg:generation}
    \begin{algorithmic}[1]
        \REQUIRE Font registry $\mathcal{F}$, text corpus $\mathcal{T}$, target count $N$, difficulty distribution $\mathcal{D}$
        \ENSURE Sample set $\mathcal{S}$
        \STATE Initialize $\mathcal{S} \leftarrow \emptyset$
        \FOR{$i = 1$ to $N$}
            \STATE Sample difficulty $d \sim \mathcal{D}$
            \STATE Sample script $s$ according to difficulty-script distribution
            \STATE Sample font $f \sim \mathcal{F}[s]$ respecting difficulty constraints
            \STATE Sample text $t \sim \mathcal{T}[s]$
            \STATE Sample size $z \in \{12, 16, 20, 24, 32, 40, 48, 64\}$
            \STATE Sample style $y \in \{\text{regular}, \text{bold}, \text{italic}, \text{bold\_italic}\}$
            \STATE Sample color $c \in \{\text{black}, \text{red}, \text{blue}, \text{green}, \ldots\}$
            \STATE Compute background color $b$ ensuring contrast with $c$
            \STATE Render image $I$ with parameters $(f, t, z, y, c, b)$
            \STATE Generate questions $Q$ for each property
            \STATE $\mathcal{S} \leftarrow \mathcal{S} \cup \{(I, Q, \text{metadata})\}$
        \ENDFOR
        \RETURN $\mathcal{S}$
    \end{algorithmic}
\end{algorithm}

\subsubsection{Font Size and Color Configuration}

Tables~\ref{tab:font_sizes_app} and~\ref{tab:colors_app} specify the eight font sizes and eight text colors used in the benchmark.

\begin{table}[h]
    \begin{minipage}[t]{0.48\textwidth}
        \centering
        \caption{Font size configuration.}
        \label{tab:font_sizes_app}
        \small
        \begin{tabular}{@{}ccc@{}}
            \toprule
            \textbf{Point Size} & \textbf{Bucket} & \textbf{Usage} \\
            \midrule
            12pt & Small & Body text \\
            16pt & Small & Captions \\
            20pt & Medium & Subheadings \\
            24pt & Medium & Subheadings \\
            32pt & Large & Headings \\
            40pt & Large & Titles \\
            48pt & XLarge & Display \\
            64pt & XLarge & Large display \\
            \bottomrule
        \end{tabular}
    \end{minipage}%
    \hfill
    \begin{minipage}[t]{0.48\textwidth}
        \centering
        \caption{Font color palette.}
        \label{tab:colors_app}
        \small
        \begin{tabular}{@{}lcc@{}}
            \toprule
            \textbf{Color} & \textbf{RGB} & \textbf{Hex} \\
            \midrule
            Black & (0, 0, 0) & \#000000 \\
            Red & (220, 50, 50) & \#DC3232 \\
            Blue & (50, 50, 220) & \#3232DC \\
            Green & (50, 150, 50) & \#329632 \\
            Gray & (128, 128, 128) & \#808080 \\
            Orange & (230, 130, 30) & \#E6821E \\
            Purple & (150, 50, 180) & \#9632B4 \\
            Brown & (140, 90, 40) & \#8C5A28 \\
            \bottomrule
        \end{tabular}
    \end{minipage}
\end{table}

\subsection{Dataset Details}
\label{app:datasets}

\subsubsection{Complete Font List}

Table~\ref{tab:complete_fonts} presents the complete list of 26 fonts included in FontBench.

\begin{table}[h]
    \centering
    \caption{Complete font list organized by script and category. All fonts are available on macOS and selected for broad typographic diversity.}
    \label{tab:complete_fonts}
    \begin{tabular}{@{}llp{10cm}@{}}
        \toprule
        \textbf{Script} & \textbf{Category} & \textbf{Fonts} \\
        \midrule
        \multirow{4}{*}{Latin (20)} & Serif & Times New Roman, Georgia, Palatino, Baskerville, Didot, Cochin, Rockwell \\
        & Sans-Serif & Arial, Arial Bold, Helvetica, Helvetica Neue, Verdana, Futura, Avenir, Gill Sans, Optima, Trebuchet MS \\
        & Monospace & Courier New, Menlo \\
        & Slab Serif & American Typewriter \\
        \midrule
        CJK (2) & Chinese & STHeiti, Songti SC \\
        \midrule
        \multirow{2}{*}{Other (4)} & Arabic & Al Nile, Baghdad \\
        & Devanagari & Devanagari MT, Devanagari Sangam MN \\
        \bottomrule
    \end{tabular}
\end{table}

\subsubsection{Text Corpus}

We use script-appropriate text for each writing system. Latin samples consist of pangrams, common phrases, and typographic test sentences in English (e.g., ``The quick brown fox jumps over the lazy dog''). Chinese samples consist of common phrases in Simplified Chinese selected for glyph coverage across the two available fonts. Arabic and Devanagari samples use representative phrases chosen for natural appearance and good glyph coverage.

\subsubsection{Sample Distribution}

Table~\ref{tab:sample_distribution} provides the complete sample distribution by script and difficulty level.

\begin{table}[h]
    \centering
    \caption{Sample distribution by script and difficulty.}
    \label{tab:sample_distribution}
    \begin{tabular}{@{}lccc@{\hspace{1em}}c@{}}
        \toprule
        & \textbf{Easy} & \textbf{Medium} & \textbf{Hard} & \textbf{Total} \\
        \midrule
        Latin & 73 & 57 & 73 & 203 \\
        CJK & 7 & 7 & 4 & 18 \\
        Other & 8 & 13 & 8 & 29 \\
        \midrule
        \textbf{Total} & 88 & 77 & 85 & 250 \\
        \bottomrule
    \end{tabular}
\end{table}

\subsection{Question Templates}
\label{app:questions}

Complete question phrasings used for each property:

\paragraph{Font Family.} ``What font family is used in this image?'' / ``Identify the typeface shown in this text.'' / ``Which font is used to render this text?'' / ``What is the name of this font?''

\paragraph{Font Size.} ``What is the approximate font size in this image?'' / ``Estimate the point size of the displayed text.'' / ``How large is the font in this image?'' / ``What size is this text rendered at?''

\paragraph{Font Style.} ``What style is the text rendered in?'' / ``Is this text regular, bold, italic, or bold-italic?'' / ``Identify the font style used here.'' / ``What typographic style is applied to this text?''

\paragraph{Font Color.} ``What color is the text in this image?'' / ``Identify the font color.'' / ``What is the color of the displayed text?'' / ``Which color is used for this text?''

\section{Evaluation Setup}
\label{app:evaluation}

\subsection{Model List}
\label{app:model_list}

Table~\ref{tab:models} summarizes all 15 VLMs evaluated in this work. The selection covers ten open-source models from four providers and five closed-source commercial APIs, spanning parameter counts from 7B to 72B and including both dense and mixture-of-experts architectures.

\begin{table}[h]
    \centering
    \caption{Summary of evaluated Vision-Language Models. Models span different sizes, architectures, and providers.}
    \label{tab:models}
    \small
    \begin{tabular}{@{}llccc@{}}
        \toprule
        \textbf{Model} & \textbf{Provider} & \textbf{Type} & \textbf{Parameters} & \textbf{Access} \\
        \midrule
        \multicolumn{5}{l}{\textit{Open-Source Models}} \\
        Qwen2.5-VL-7B & Alibaba & Dense & 7B & Weights \\
        Qwen2.5-VL-32B & Alibaba & Dense & 32B & Weights \\
        Qwen2.5-VL-72B & Alibaba & Dense & 72B & Weights \\
        Qwen3-VL-8B & Alibaba & Dense & 8B & Weights \\
        Qwen3-VL-30B-A3B & Alibaba & MoE & 30B (3B active) & Weights \\
        Qwen3-VL-32B & Alibaba & Dense & 32B & Weights \\
        Pixtral-12B & Mistral & Dense & 12B & Weights \\
        GLM-4.5V & Zhipu AI & MoE & 108B (12B active) & Weights \\
        GLM-4.6V & Zhipu AI & Dense & 106B & Weights \\
        \midrule
        \multicolumn{5}{l}{\textit{Closed-Source Models}} \\
        Qwen3-Max & Alibaba & Proprietary & Unknown & API \\
        GPT-5.2 & OpenAI & Proprietary & Unknown & API \\
        Gemini-3-Flash & Google & Proprietary & Unknown & API \\
        Gemini-3-Pro & Google & Proprietary & Unknown & API \\
        Claude-Sonnet-4.6 & Anthropic & Proprietary & Unknown & API \\
        Doubao-Seed-1.6 & ByteDance & Proprietary & Unknown & API \\
        \bottomrule
    \end{tabular}
\end{table}

\subsection{Implementation Details}
\label{app:implementation}

\subsubsection{Image Rendering}

Images are rendered using Python's Pillow library. Each image is PNG-format RGB with anti-aliasing enabled, 96 DPI, and a background color sampled to ensure a minimum contrast ratio of 4.5:1 with the text color. Canvas size is dynamically computed from text bounding-box bounds with 20 px padding on all sides.

\subsubsection{API Configuration}

All models are accessed with temperature 0, \texttt{max\_tokens} 100, and default sampling parameters (\texttt{top\_p}~1.0, no frequency or presence penalties).

\subsubsection{Prompt Template}

\begin{tcolorbox}[colback=gray!5, colframe=gray!50, title=Evaluation Prompt Template, breakable]
\small\ttfamily
Look at this image and answer the following\\
question.\\[0.5em]
Question: \{question\}\\[0.5em]
Options:\\
A. \{option\_a\}\\
B. \{option\_b\}\\
C. \{option\_c\}\\
D. \{option\_d\}\\[0.5em]
Respond with only the letter (A, B, C, or D)\\
of your answer.
\end{tcolorbox}

\subsubsection{Answer Parsing}

Responses are parsed with the following priority cascade: (1) exact letter match (``A'', ``B'', ``C'', ``D''); (2) letter with surrounding punctuation (``A.'', ``(A)'', ``A)''); (3) first valid letter appearing in the response; (4) full option-text match. Responses that fail all four steps are marked incorrect.

\subsection{Compute Resources}
\label{app:compute}

All evaluations consist of 1{,}000 API calls per model (250 samples $\times$ 4 questions). Total wall-clock time ranged from approximately 10 minutes (Qwen2.5-VL-7B via local inference) to 161 minutes (GLM-4.5V) depending on API latency. No GPU is required for evaluation; all inference runs on the respective model providers' infrastructure.

\section{Complete Results and Statistical Analysis}
\label{app:results}

\subsection{Complete Results}
\label{app:additional_results}

Table~\ref{tab:full_results} reports per-property and per-difficulty accuracy for all evaluated models.

\begin{longtable}{@{}lcccc@{\hspace{1em}}cccc@{}}
    \caption{Accuracy (\%) by property and difficulty for all models.}
    \label{tab:full_results} \\
    \toprule
    \multirow{2}{*}{\textbf{Model}} & \multicolumn{4}{c}{\textbf{By Property}} & \multicolumn{4}{c}{\textbf{By Difficulty}} \\
    \cmidrule(lr){2-5} \cmidrule(lr){6-9}
    & Family & Size & Style & Color & Easy & Med & Hard & Overall \\
    \midrule
    \endfirsthead
    \multicolumn{9}{c}{\tablename\ \thetable{} -- continued from previous page} \\
    \toprule
    \multirow{2}{*}{\textbf{Model}} & \multicolumn{4}{c}{\textbf{By Property}} & \multicolumn{4}{c}{\textbf{By Difficulty}} \\
    \cmidrule(lr){2-5} \cmidrule(lr){6-9}
    & Family & Size & Style & Color & Easy & Med & Hard & Overall \\
    \midrule
    \endhead
    \midrule
    \multicolumn{9}{r}{Continued on next page} \\
    \endfoot
    \bottomrule
    \endlastfoot
    \multicolumn{9}{l}{\textit{Open-Source Models}} \\
    Qwen2.5-VL-7B & 35.2 & 44.4 & 27.6 & 97.6 & 51.4 & 53.2 & 49.1 & 51.2 \\
    Qwen2.5-VL-32B & 36.0 & 34.4 & 30.4 & 89.2 & 49.4 & 45.1 & 47.6 & 47.5 \\
    Qwen2.5-VL-72B & 38.8 & 36.0 & 33.2 & 96.4 & 49.7 & 52.6 & 51.2 & 51.1 \\
    Qwen3-VL-8B & 36.8 & 39.2 & 28.8 & 99.6 & 50.3 & 53.2 & 50.0 & 51.1 \\
    Qwen3-VL-30B-A3B & 49.6 & 40.4 & 28.0 & 99.6 & 54.3 & 54.9 & 54.1 & 54.4 \\
    Qwen3-VL-32B & 42.4 & 37.2 & 26.0 & 100.0 & 51.4 & 52.9 & 50.0 & 51.4 \\
    Pixtral-12B & 26.0 & 27.2 & 28.4 & 24.8 & 25.9 & 26.6 & 27.4 & 26.6 \\
    GLM-4.5V & 25.2 & 22.0 & 22.8 & 26.8 & 22.7 & 24.0 & 25.9 & 24.2 \\
    GLM-4.6V & 39.2 & 35.6 & 25.6 & 100.0 & 48.3 & 50.3 & 51.8 & 50.1 \\
    \midrule
    \multicolumn{9}{l}{\textit{Closed-Source Models}} \\
    Qwen3-Max & 46.0 & 32.4 & 31.2 & 99.2 & 51.4 & 52.6 & 52.6 & 52.2 \\
    GPT-5.2 & 58.8 & 50.0 & 31.2 & 99.6 & 61.6 & 59.7 & 58.2 & 59.9 \\
    Gemini-3-Flash & 80.8 & 52.4 & 33.6 & 100.0 & 65.9 & 67.2 & 67.1 & 66.7 \\
    Gemini-3-Pro & 40.8 & 41.2 & 32.0 & 94.0 & 50.9 & 53.2 & 52.1 & 52.0 \\
    Claude-Sonnet-4.6 & 64.0 & 44.8 & 28.0 & 97.6 & 60.2 & 57.8 & 57.6 & 58.6 \\
    Doubao-Seed-1.6 & 44.8 & 44.4 & 30.4 & 98.8 & 53.1 & 57.8 & 53.2 & 54.6 \\
\end{longtable}

\subsection{Statistical Analysis}
\label{app:statistics}

\subsubsection{Confidence Intervals}

We compute 95\% confidence intervals using the Wilson score interval for binomial proportions ($n{=}1000$ questions per model).

\begin{table}[h]
    \centering
    \caption{Overall accuracy with 95\% confidence intervals ($n{=}1000$ questions per model).}
    \label{tab:confidence}
    \begin{tabular}{@{}lcc@{}}
        \toprule
        \textbf{Model} & \textbf{Accuracy (\%)} & \textbf{95\% CI} \\
        \midrule
        Gemini-3-Flash & 66.7 & [63.7, 69.6] \\
        GPT-5.2 & 59.9 & [56.8, 62.9] \\
        Claude-Sonnet-4.6 & 58.6 & [55.5, 61.6] \\
        Doubao-Seed-1.6 & 54.6 & [51.5, 57.7] \\
        Qwen3-VL-30B-A3B & 54.4 & [51.3, 57.5] \\
        Qwen3-Max & 52.2 & [49.1, 55.3] \\
        Gemini-3-Pro & 52.0 & [48.9, 55.1] \\
        Qwen3-VL-32B & 51.4 & [48.3, 54.5] \\
        Qwen2.5-VL-7B & 51.2 & [48.1, 54.3] \\
        Qwen2.5-VL-72B & 51.1 & [48.0, 54.2] \\
        Qwen3-VL-8B & 51.1 & [48.0, 54.2] \\
        GLM-4.6V & 50.1 & [47.0, 53.2] \\
        Qwen2.5-VL-32B & 47.5 & [44.4, 50.6] \\
        Pixtral-12B & 26.6 & [23.9, 29.5] \\
        GLM-4.5V & 24.2 & [21.6, 27.0] \\
        \bottomrule
    \end{tabular}
\end{table}

\subsubsection{Statistical Significance}

Using McNemar's test for pairwise model comparisons: Gemini-3-Flash is significantly better than all other models ($p < 0.001$). GPT-5.2 and Claude-Sonnet-4.6 are significantly better than middle-tier models ($p < 0.05$). Pairwise differences among the middle-tier cluster (Qwen3-VL-30B-A3B, Doubao-Seed-1.6, Qwen3-Max, Gemini-3-Pro, Qwen3-VL-32B, Qwen2.5-VL-7B, Qwen2.5-VL-72B, Qwen3-VL-8B) are not statistically significant ($p > 0.05$), consistent with their overlapping confidence intervals. GLM-4.5V and Pixtral-12B are not significantly different from the 25\% random baseline ($p > 0.05$).

\subsection{Perception Hierarchy Illustration}
\label{app:hierarchy}

\begin{figure}[h]
    \centering
    \includegraphics[width=0.85\textwidth]{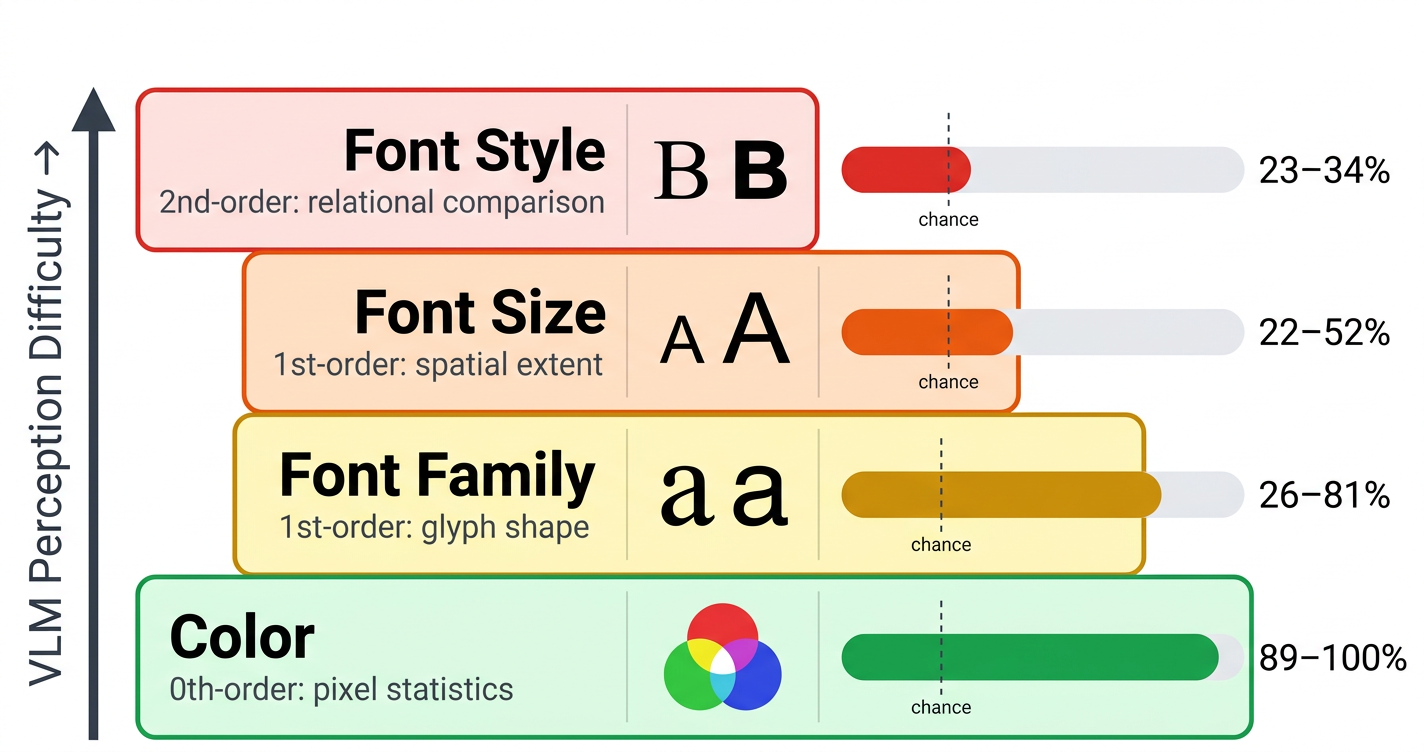}
    \caption{\textbf{Perception hierarchy.} Typographic properties map onto increasing visual feature complexity: color (0th-order pixel statistics, 89--100\%) $\gg$ family (1st-order glyph shape, 26--81\%) $>$ size (1st-order spatial extent, 22--52\%) $>$ style (2nd-order relational comparison, 23--34\%). Dashed line marks the 25\% random-chance baseline.}
    \label{fig:hierarchy}
\end{figure}

Figure~\ref{fig:hierarchy} visualizes the perception hierarchy discussed in \S\ref{subsec:hierarchy} of the main paper. The four typographic properties are arranged by their computational complexity within ViT-based encoders, from zeroth-order pixel statistics for color to second-order relational comparisons for style.

\subsection{Accuracy Heatmap}
\label{app:heatmap}

\begin{figure}[h]
    \centering
    \includegraphics[width=0.75\textwidth]{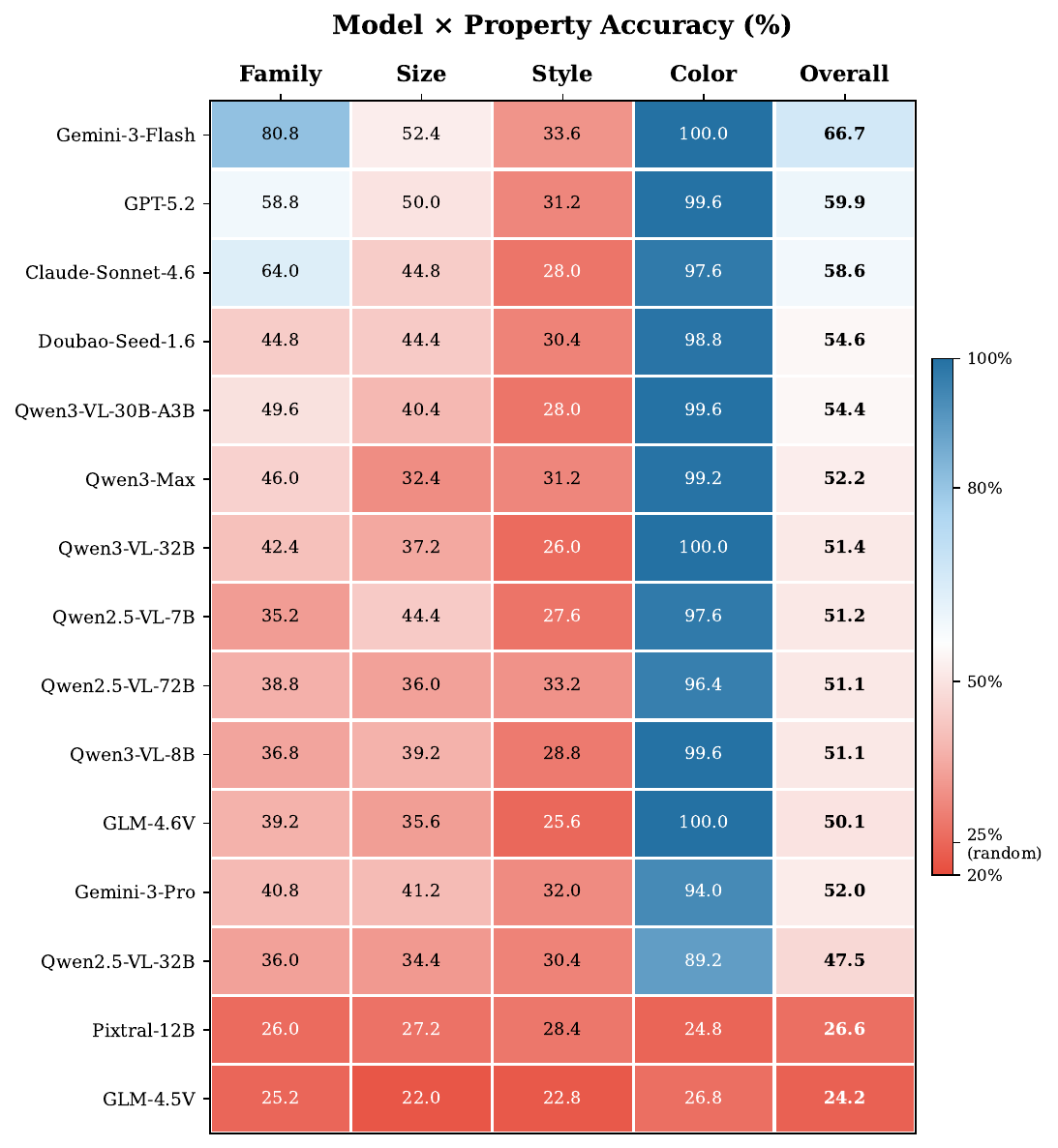}
    \caption{Model $\times$ property accuracy heatmap. A clear perception hierarchy emerges: color is trivially solved, while style remains universally poor. Gemini-3-Flash leads across all properties; GLM-4.5V performs at near-random levels.}
    \label{fig:heatmap}
\end{figure}

Figure~\ref{fig:heatmap} presents a per-model, per-property accuracy heatmap that provides a comprehensive view of the results in Table~2 of the main paper. Models are sorted by overall accuracy from top to bottom.

\section{Fine-Tuning Details}
\label{app:finetuning}

\subsection{LoRA Configuration}
\label{app:lora}

\begin{lstlisting}[language=Python, basicstyle=\small\ttfamily]
lora_config = {
    "r": 16,
    "lora_alpha": 32,
    "lora_dropout": 0.05,
    "target_modules": ["q_proj", "k_proj", "v_proj", "o_proj"],
    "task_type": "CAUSAL_LM"
}

training_args = {
    "learning_rate": 2e-4,
    "num_epochs": 3,
    "batch_size": 1,
    "gradient_accumulation_steps": 16,
    "warmup_ratio": 0.05,
    "bf16": True,
    "gradient_checkpointing": True
}
\end{lstlisting}

\subsection{Training Data}
\label{app:training_data}

Fine-tuning uses 3{,}000 image-question-answer triplets generated by the same synthetic pipeline described in \S\ref{sec:method} of the main paper, formatted as instruction-tuning conversations compatible with each model's chat template. The training set covers all four font properties with balanced distributions across difficulty levels and writing systems. Training images use different random seeds and text samples than the evaluation benchmark to prevent data leakage.

\section{Cross-Benchmark Validation}
\label{app:crossbenchmark}

\subsection{Full FRB Cross-Benchmark Results}
\label{app:frb_full}

Table~\ref{tab:frb_results} reports the complete cross-benchmark evaluation on FRB, including results for all 15 baseline models and three fine-tuned variants under both easy and hard (stroop) conditions.

\begin{table}[h]
    \centering
    \caption{Cross-benchmark evaluation on the FRB 15-font MCQ format~\citep{li2025texture}. Easy: normal sentences; Hard: stroop condition where the text spells a font name. Random baseline is 6.7\%.}
    \label{tab:frb_results}
    \small\setlength{\tabcolsep}{5pt}
    \begin{tabular}{@{}lccc@{}}
        \toprule
        \textbf{Model} & \textbf{FRB Easy} & \textbf{FRB Hard} & \textbf{FRB Overall} \\
        \midrule
        \multicolumn{4}{l}{\textit{Open-Source Models}} \\
        Qwen2.5-VL-7B & 7.3 & 6.7 & 6.9 \\
        Qwen2.5-VL-32B & 8.7 & 6.7 & 7.5 \\
        Qwen2.5-VL-72B & 9.3 & 6.7 & 7.7 \\
        Qwen3-VL-8B & 14.0 & 6.7 & 9.6 \\
        Qwen3-VL-30B-A3B & 14.7 & 6.7 & 9.9 \\
        Qwen3-VL-32B & 14.0 & 6.7 & 9.6 \\
        Pixtral-12B & 9.3 & 8.4 & 8.8 \\
        GLM-4.5V & 2.7 & 7.1 & 5.3 \\
        GLM-4.6V & 9.3 & 6.7 & 7.7 \\
        \midrule
        \multicolumn{4}{l}{\textit{Closed-Source Models}} \\
        Qwen3-Max & 15.3 & 6.7 & 10.1 \\
        GPT-5.2 & 26.7 & 16.9 & 20.8 \\
        Gemini-3-Flash & 56.7 & 29.8 & 40.5 \\
        Gemini-3-Pro & 24.7 & 10.7 & 16.3 \\
        Claude-Sonnet-4.6 & 34.0 & 15.6 & 22.9 \\
        Doubao-Seed-1.6 & 10.7 & 6.7 & 8.3 \\
        \midrule
        \multicolumn{4}{l}{\textit{Fine-Tuned Models}} \\
        Qwen2.5-VL-7B + LoRA & 14.0 & 6.7 & 9.6 \\
        Qwen3-VL-8B + LoRA & 22.7 & 6.7 & 13.1 \\
        Qwen2.5-VL-32B + LoRA & 23.3 & 6.7 & 13.3 \\
        \midrule
        Random Baseline & 6.7 & 6.7 & 6.7 \\
        \bottomrule
    \end{tabular}
\end{table}

\subsection{Stroop Effect Visualization}
\label{app:stroop}

\begin{figure}[h]
    \centering
    \includegraphics[width=\textwidth]{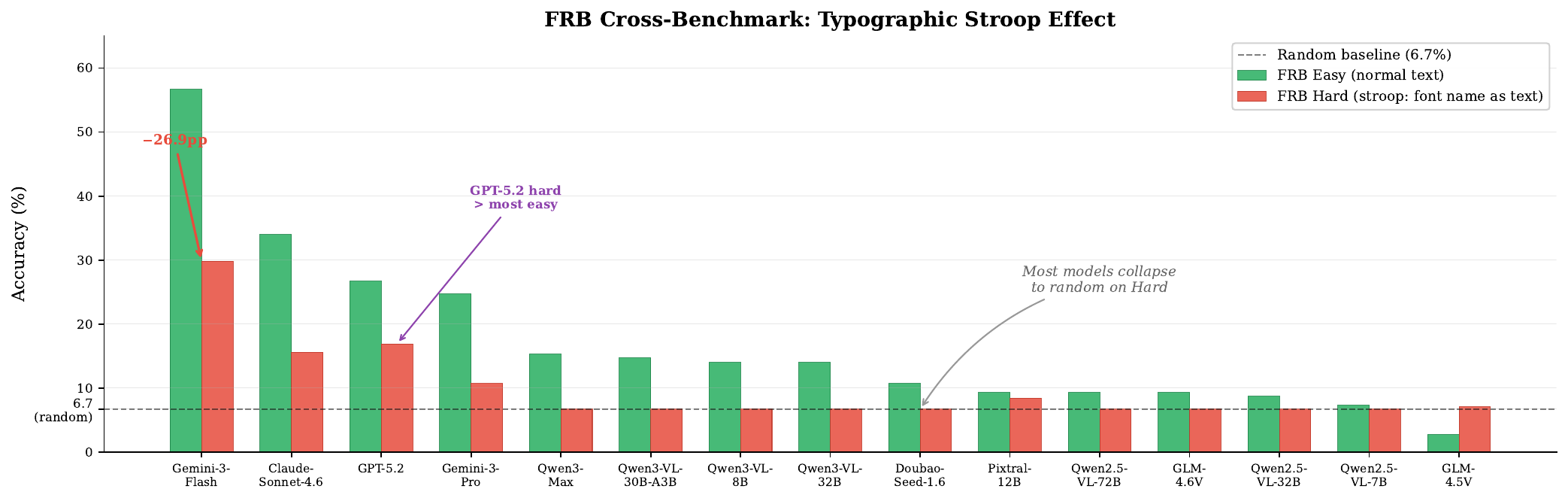}
    \caption{The stroop effect~\citep{li2025texture} evaluated across all models. Under normal text, Gemini-3-Flash achieves 56.7\%, but when text spells a conflicting font name, most models collapse to the 6.7\% random baseline. Claude-Sonnet-4.6 drops from 34.0\% to 15.6\%.}
    \label{fig:stroop}
\end{figure}

Figure~\ref{fig:stroop} visualizes the stroop effect across all models, comparing FRB-easy and FRB-hard accuracy side by side. The dramatic drop under the stroop condition reveals the extent to which VLMs rely on text content rather than visual features for font identification.

\section{Detailed Analysis}
\label{app:analysis}

\subsection{Per-Property Detailed Analysis}
\label{app:property_analysis}

\subsubsection{Font Family}

Font family recognition varies widely across models, with Gemini-3-Flash achieving 80.8\% while most open-source models hover around 35--50\%. The difficulty stems from subtle visual differences between fonts; distinguishing Arial from Helvetica requires perceiving minute variations in character proportions and stroke terminals. Three failure patterns dominate. \emph{Within-category confusion}: models frequently mistake one sans-serif font for another but rarely confuse a serif font with a monospace one. \emph{Default-response bias}: some models overwhelmingly predict common fonts like Arial or Times New Roman regardless of actual content. \emph{OCR interference}: models appear to focus on reading the text rather than perceiving its visual presentation.

\subsubsection{Font Size}

Table~\ref{tab:size_buckets} shows performance by size bucket. Models exhibit strikingly different size perception strategies. Qwen3-VL-8B and Gemini-3-Flash both achieve 0\% on XLarge sizes (48--64pt), systematically predicting ``large'' instead, suggesting these models lack the concept of extra-large font sizes. In contrast, GPT-5.2 achieves 90.3\% on XLarge and 76.3\% on Small, demonstrating accurate perception at size extremes but struggling with intermediate sizes at only 35.5\% on Medium.

\begin{table}[h]
    \centering
    \begin{minipage}[t]{0.52\textwidth}
        \centering
        \caption{Font size recognition accuracy (\%) by size bucket.}
        \label{tab:size_buckets}
        \small
        \begin{tabular}{@{}lcccc@{}}
            \toprule
            \textbf{Model} & \textbf{S} & \textbf{M} & \textbf{L} & \textbf{XL} \\
            \midrule
            Qwen3-VL-8B & 18.6 & 58.9 & 45.3 & 0.0 \\
            Gemini-3-Flash & 57.6 & 46.7 & \best{88.7} & 0.0 \\
            GPT-5.2 & \best{76.3} & 35.5 & 45.3 & \best{90.3} \\
            \bottomrule
        \end{tabular}
    \end{minipage}%
    \hfill
    \begin{minipage}[t]{0.45\textwidth}
        \centering
        \caption{Style confusion matrix for Qwen3-VL-8B (\%).}
        \label{tab:style_confusion}
        \small
        \begin{tabular}{@{}lcccc@{}}
            \toprule
            \textbf{True\,/\,Pred} & \textbf{Reg} & \textbf{Bld} & \textbf{Ita} & \textbf{B-I} \\
            \midrule
            Regular & 69.2 & 14.1 & 16.7 & 0.0 \\
            Bold & 70.5 & 13.7 & 15.8 & 0.0 \\
            Italic & 80.0 & 5.0 & 12.5 & 2.5 \\
            Bold-Italic & 67.6 & 10.8 & 21.6 & 0.0 \\
            \bottomrule
        \end{tabular}
    \end{minipage}
\end{table}

\subsubsection{Font Style}

Style recognition remains challenging, with the best model achieving only 33.6\%. The confusion matrix in Table~\ref{tab:style_confusion} reveals a dominant ``regular'' bias: the model predicts regular for 67.6--80.0\% of \emph{all} samples regardless of true style, including 70.5\% of bold and 80.0\% of italic samples. Bold-italic is essentially never predicted (0.0\% for three of four true styles). This extreme bias suggests that VLMs largely ignore stylistic visual cues and default to the most common style, treating font style as invisible metadata rather than a perceivable visual attribute.

\subsubsection{Font Color}

Color recognition achieves near-perfect accuracy for most models, with most achieving $\geq$89\% and most exceeding 97\%. Gemini-3-Flash is flawless at 100\% across all eight colors (Table~\ref{tab:color_accuracy}). Qwen2.5-VL-32B is a notable exception at 89.2\%, with consistent errors across all colors, suggesting a systematic issue in its color pipeline rather than difficulty with specific hues.

\begin{table}[h]
    \centering
    \caption{Color recognition accuracy (\%) by color. Most models achieve near-perfect accuracy; GLM-4.5V fails across all colors.}
    \label{tab:color_accuracy}
    \small
    \begin{tabular}{@{}lcccccccc@{}}
        \toprule
        \textbf{Model} & \textbf{Black} & \textbf{Red} & \textbf{Blue} & \textbf{Green} & \textbf{Gray} & \textbf{Orange} & \textbf{Purple} & \textbf{Brown} \\
        \midrule
        Qwen3-VL-8B & 100 & 100 & 97 & 100 & 100 & 100 & 100 & 100 \\
        Gemini-3-Flash & 100 & 100 & 100 & 100 & 100 & 100 & 100 & 100 \\
        Doubao-Seed-1.6 & 100 & 100 & 100 & 100 & 97 & 100 & 100 & 93 \\
        GLM-4.5V & 28 & 40 & 21 & 31 & 19 & 32 & 23 & 19 \\
        \bottomrule
    \end{tabular}
\end{table}

\subsection{Scaling Paradox Details}
\label{app:scaling}

\begin{table}[h]
    \centering
    \caption{Model size vs.\ overall accuracy. Larger models do not consistently outperform smaller ones.}
    \label{tab:size_comparison}
    \begin{tabular}{@{}lcc@{}}
        \toprule
        \textbf{Model Comparison} & \textbf{Smaller} & \textbf{Larger} \\
        \midrule
        Qwen2.5-VL-7B vs 32B & 51.2\% & 47.5\% \\
        Qwen2.5-VL-32B vs 72B & 47.5\% & 51.1\% \\
        Qwen3-VL-8B vs 30B-A3B & 51.1\% & 54.4\% \\
        Qwen3-VL-8B vs 32B & 51.1\% & 51.4\% \\
        \bottomrule
    \end{tabular}
\end{table}

In the Qwen2.5 family, the scaling curve is non-monotonic: the 7B model at 51.2\% outperforms the 32B model at 47.5\% by 3.7 pp, while the 72B model recovers only to 51.1\%, nearly matching the 7B. The 32B model shows notably degraded color recognition (89.2\% vs 97.6\% for the 7B) and worse font size accuracy (34.4\% vs 44.4\%), suggesting that intermediate scales can introduce regressions in fine-grained visual perception. In the Qwen3 family, the 30B-A3B MoE variant at 54.4\% outperforms both the 8B at 51.1\% and the 32B dense model at 51.4\%, with improvements concentrated in font family recognition (49.6\% vs 36.8\% for the 8B).

\subsection{Error Analysis}
\label{app:error_analysis}

Several systematic failure modes emerge across models. The most pervasive is a \emph{default-font bias}: many models overwhelmingly predict common typefaces like Arial or Times New Roman regardless of actual content, reflecting strong prior biases from web-scale pretraining corpora where these fonts dominate. A subtler failure mode arises when models reason semantically rather than visually: we observe responses such as ``this appears to be formal text, likely Times New Roman,'' where the model infers font identity from text content rather than visual evidence. This OCR-then-reason shortcut is a fundamental limitation, suggesting that font-related queries are routed through the language modeling pathway rather than the visual encoder. Some models also exhibit positional bias, favoring certain answer positions when visual perception fails to produce a confident prediction; our randomized option ordering mitigates this on aggregate metrics but does not eliminate it.

To place model performance in perspective, prior work on font identification shows that untrained individuals can distinguish broad typeface categories such as serif and sans-serif reliably, but struggle with within-category identification~\citep{wang2015deepfont}. Professional designers, through years of training, develop substantially sharper font discrimination~\citep{garfield2012fonts}. The current best VLM, Gemini-3-Flash at 80.8\% font family accuracy, performs comparably to an untrained human attempting within-category font discrimination, well below the level of a trained typographer. For color recognition, most VLMs achieve near-ceiling accuracy of 99--100\%, matching or exceeding typical human response accuracy under controlled conditions~\citep{berlin1991basic}.

\subsection{Failure Case Analysis}
\label{app:failures}

\paragraph{Within-category confusion.} Models frequently confuse fonts within the same category: Arial, Helvetica, and Verdana are mutually confused among sans-serif fonts; Times New Roman, Georgia, and Palatino among serif fonts; and Helvetica and Helvetica Neue represent the hardest variant-level confusion.

\paragraph{Default-response collapse.} Pixtral-12B (26.6\% overall, 24.8\% color) and GLM-4.5V (24.2\% overall, 26.8\% color) perform near the 25\% random baseline across all four properties including color, indicating that these models do not engage with the image at all for typographic queries.

\paragraph{Size estimation errors.} Adjacent sizes are the most common confusion (e.g., 24 pt vs.\ 32 pt). Some models show near-zero accuracy on the XLarge bucket (48--64 pt), suggesting they never output the extreme size labels. Middle-range sizes (20--40 pt) are the hardest for most models due to overlapping relative appearance.

\section{Discussion}
\label{app:discussion}

\subsection{Limitations}
\label{app:limitations}

\textbf{Benchmark scale.} FontBench comprises 250 samples yielding 1{,}000 questions. While sufficient to detect large performance differences (95\% confidence intervals span $\pm$3\,pp), this scale limits statistical power for fine-grained comparisons. Non-Latin coverage is particularly sparse: CJK has only 18 samples and Arabic only 11.

\textbf{Synthetic data.} All evaluation images are synthetically rendered with clean backgrounds and controlled typography. This enables precise annotation but does not capture real-world variability: mixed fonts, degraded printing, and complex layouts. Performance on FontBench may overestimate real-world typographic perception.

\textbf{Property independence.} Each property is evaluated in isolation. Real-world typography perception often involves joint reasoning (e.g., heading detection requires perceiving both larger size and bold style simultaneously). Multi-attribute interactions remain unexplored.

\textbf{Fine-tuning scope.} LoRA experiments are limited to three Qwen-family models; generalization to other architectures remains untested. The 32B model's quantization confound prevents clean scaling conclusions.

\subsection{Future Directions}
\label{app:future}

For model developers, the perception hierarchy provides a concrete roadmap: color is solved, family and size are learnable with modest synthetic data, and style demands new architectural approaches for relational visual reasoning. The capability-fragility trade-off further warns that models with the highest clean-image accuracy may be the least reliable under blur or compression. For practitioners in typography-sensitive domains such as document analysis, accessibility tools, and design automation, our fine-tuning recipe offers an immediate and resource-efficient path to improvement.

Key directions include extending FontBench to real-world document images, broadening non-Latin script and font coverage, and investigating contrastive or relational architectures for style perception. On the broader front, reliable font recognition supports beneficial applications in accessibility, cultural preservation, and design automation. We note that improved capabilities could also facilitate document forgery and encourage responsible development.

\subsection{Reproducibility}
\label{app:reproducibility}

All random seeds are fixed at 42. API calls use temperature 0 for deterministic responses. The benchmark dataset, evaluation code, and detailed per-model result logs will be publicly released upon acceptance.

\end{document}